\documentclass[11pt]{article}

\usepackage[final]{acl}

\usepackage{times}
\usepackage{latexsym}

\usepackage[T1]{fontenc}

\usepackage[utf8]{inputenc}

\usepackage{microtype}

\usepackage{inconsolata}

\newcommand{\ie}{\textit{i.e.}}
\newcommand{\eg}{\textit{e.g.}}

\usepackage{amsmath,amsfonts,bm}

\def\eqref#1{equation~\ref{#1}}
\def\1{\bm{1}}

\def\vx{{\bm{x}}}

\def\mA{{\bm{A}}}

\def\mK{{\bm{K}}}

\def\mM{{\bm{M}}}

\def\mO{{\bm{O}}}

\def\mQ{{\bm{Q}}}
\def\mR{{\bm{R}}}

\def\mV{{\bm{V}}}

\def\mX{{\bm{X}}}

\DeclareMathAlphabet{\mathsfit}{\encodingdefault}{\sfdefault}{m}{sl}
\SetMathAlphabet{\mathsfit}{bold}{\encodingdefault}{\sfdefault}{bx}{n}

\usepackage{graphicx}
\usepackage{subcaption}
\usepackage{booktabs}
\usepackage[table]{xcolor}
\usepackage{multirow}
\usepackage{wrapfig}
\usepackage{array}
\usepackage{makecell}
\usepackage{float}

\usepackage{amsmath}
\usepackage{amssymb}
\usepackage{mathtools}
\usepackage{algorithm}
\usepackage{algorithmic}
\usepackage{xcolor}
\newcommand{\method}{SinkPruner}

\definecolor{revcolor}{RGB}{0,90,160}
\newcommand{\rev}[1]{#1}
\title{\method{}: Sink-Free Visual Token Pruning \\for Multimodal Large Language Models}

\newcommand\blfootnote[1]{%
  \begingroup
  \renewcommand\thefootnote{}\footnote{#1}%
  \addtocounter{footnote}{-1}%
  \endgroup
}

\author{
Shiyu Li$^{*1}$,
Zi-Yuan Hu$^{*1}$,
Shijia Huang$^2$, 
Yanyang Li$^2$,
Yiwu Zhong$^3$, 
Liwei Wang$^{\# 1}$\\
$^1$The Chinese University of Hong Kong \quad 
$^2$Weitu AI \quad
$^3$Peking University \\
}

\begin{document}
\maketitle

% First-page author note: affiliations, merged e-mail addresses, equal
% contribution and correspondence. Rendered without a footnote number.
% \authorfootnote{%
%   \textsuperscript{*}Equal contribution.\quad
%   \textsuperscript{1}Department of Computer Science and Engineering,
%   The Chinese University of Hong Kong.\quad
%   \textsuperscript{2}School of Intelligence Science and Technology,
%   Peking University.\quad
%   E-mail: \texttt{\{syli25, zyhu22, sjhuang, yyli21, lwwang\}@cse.cuhk.edu.hk},
%   \texttt{yiwu-zhong@outlook.com}.\quad
%   \textsuperscript{\textdagger}Corresponding author: Liwei Wang
%   \textless\href{mailto:lwwang@cse.cuhk.edu.hk}{lwwang@cse.cuhk.edu.hk}\textgreater.
% }

% -----------------------------------------------------------------------------
% Main paper sections migrated from ECCV main.tex
% -----------------------------------------------------------------------------
% sec/0_abstract is assumed to already contain:
% \begin{abstract}
% ...
% \end{abstract}

\begin{abstract}

Despite their strong multimodal understanding ability, multimodal large language models (MLLMs) incur substantial computational overhead when processing long visual token sequences.
To reduce inference costs, recent studies have explored visual token pruning through vision-centric or text-guided strategies.
However, these methods often overlook high-norm outlier tokens, \ie, tokens with abnormally large feature norms, leading to suboptimal pruning decisions.
In this work, we show that such high-norm outlier tokens are highly redundant in both feature and spatial dimensions, yet are often mistakenly preserved as informative cues by existing methods.
Motivated by this observation, we propose \textbf{\method{}}, a training-free visual token pruning framework for efficient MLLM inference.
\textit{\method{}} follows a coarse-to-fine design with two key modules: a \textbf{visual sanitizer} that filters high-norm redundancies and alleviates attention sink and attention dispersion, and a \textbf{text-guided pruner} that further retains tokens semantically aligned with the text query.
Extensive experiments on twelve image-language and four video-language benchmarks demonstrate the effectiveness, efficiency, and generalizability of our framework.
Notably, \textit{\method{}} preserves 96.5\% (91.8\%) of the original performance of LLaVA-1.5 (Qwen2.5-VL) under an 89\% token reduction.
Experiments further indicate that our \textit{visual sanitizer} exhibits promising transferability in enhancing the performance of existing pruning methods.
Our code is available at \href{https://github.com/LaVi-Lab/SinkPruner}{https://github.com/LaVi-Lab/SinkPruner}.
\blfootnote{$^*$equal contributions, $^\#$ corresponding author.}

% \keywords{Visual Token Pruning \and Efficient MLLM Inference}
\end{abstract}
\section{Introduction}
\label{sec:intro}

% [to add a radar chart as Fig1? refer to visionzip]
\begin{figure}[t]
    \centering

    % -------------------- (a) top --------------------
    \begin{subfigure}[t]{0.9\linewidth}
        \centering
        \includegraphics[height=3.5cm]{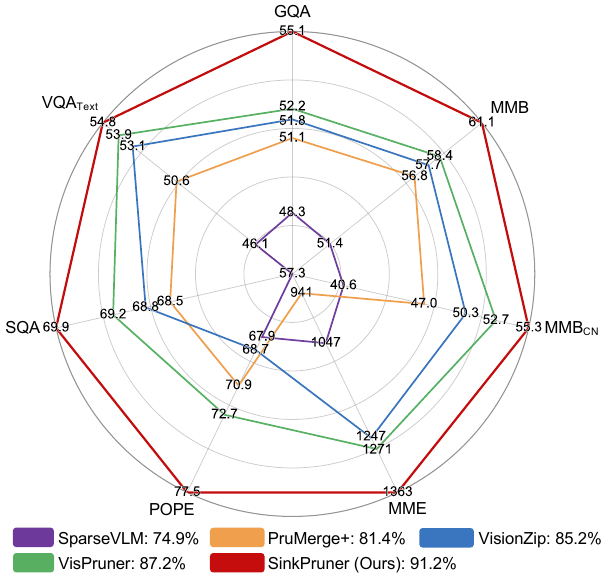}
        \caption{Performance}
        \label{fig:radar}
    \end{subfigure}

    \vspace{0.4em}

    % -------------------- (b)(c) bottom --------------------
    \begin{subfigure}[t]{0.47\linewidth}
        \centering
        \includegraphics[height=3cm]{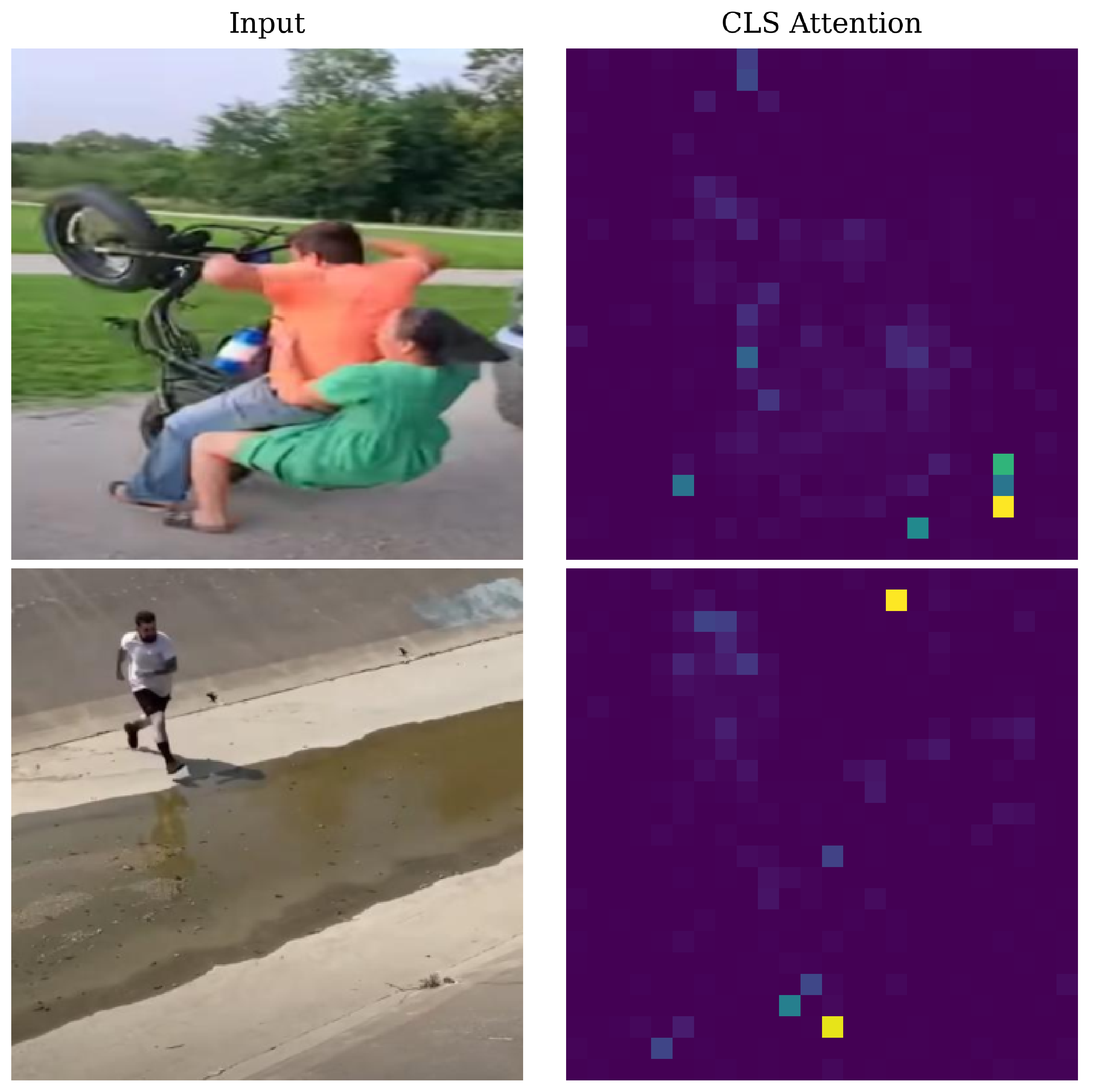}
        \caption{Attention Outliers}
        \label{fig:attn_artifacts}
    \end{subfigure}
    \hspace{0.1em}
    \begin{subfigure}[t]{0.47\linewidth}
        \centering
        \includegraphics[height=3cm]{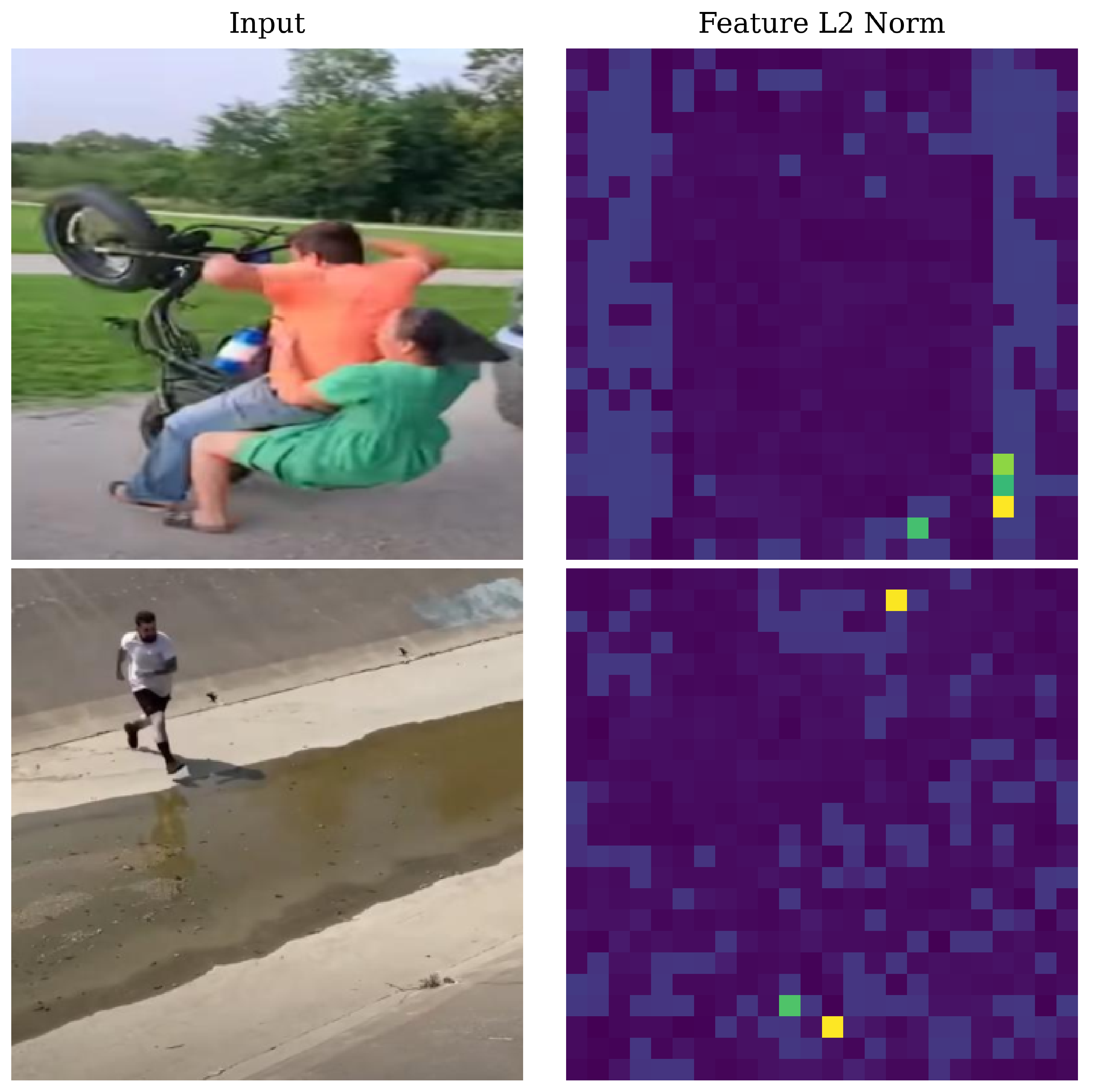}
        \caption{$L_{2}$ Norm}
        \label{fig:l2_norm}
    \end{subfigure}
    
    \caption{
        \textbf{Motivation and Performance Overview.}
        (a) \textit{\method{}} achieves better performance trade-offs than SOTA methods.
        (b) Attention maps reveal that peaky outliers (bright spots) often appear in background regions, distracting the model.
        (c) Visual feature $L_2$ norm heatmap. These attention outliers spatially align with tokens of abnormally high norms (bright spots), confirming that high attention is driven by \textbf{high-norm outliers} rather than semantic importance.
    }
    \label{fig:teaser_combined}
    \vspace{-20pt}
\end{figure}

Recent advances in large language models (LLMs)~\cite{gpt4_tech, qwen_tech, llama_tech} have catalyzed a new generation of multimodal LLMs (MLLMs)~\cite{Mini_Gemini24, qwen_vl23, blip_23, llava1_5}, achieving remarkable capabilities in both image and video understanding~\cite{vqa1, vqa2, vqa3, video1, video2}. 
To align visual modalities with textual semantics, MLLMs typically project images or videos into extensive sequences of visual tokens, which are then jointly processed by the LLM decoder~\cite{llava1_5, qwen25, llama_tech}. 
However, this paradigm suffers from a practical bottleneck: the quadratic computational complexity of Transformer attention mechanisms~\cite{vaswani2017attention, vit2020}, compounded by excessively long visual sequences, incurs prohibitive computational and memory costs during MLLM inference. 
This inefficiency restricts the context window availability for textual reasoning and critically hinders deployment in resource-constrained environments, such as edge computing and robotics~\cite{sharshar2025vision,edge1,edge2,edge3}.

To alleviate the token burden, prior studies have explored visual token pruning and compression methods for efficient MLLM inference, which can be broadly categorized into {vision-centric} and {text-guided} strategies. 
Specifically, {vision-centric} methods (\eg, VisionZip~\cite{VisionZip25}, HoloV~\cite{holov25}) operate within the vision encoder and remove redundancy based on visual cues, such as \textit{CLS attention} or feature similarity among visual tokens. 
{Text-guided} methods (\eg, FastV~\cite{FastV24}, SparseVLM~\cite{SparseVLM25}) leverage the LLM decoder to select tokens that appear semantically aligned with the user query, typically via text-visual attention. 

Despite their progress, existing pruning methods suffer from critical limitations rooted in two pathological phenomena within the MLLM pipeline. 
First, attention-based pruning at the vision encoder is misled by high-norm outliers, \ie, outlier tokens characterized by abnormally large feature norms. State-of-the-art methods heavily rely on \textit{CLS attention} scores~\cite{VisionZip25, vispruner25, FasterVLM24} to retain important tokens. However, this criterion systematically prioritizes \emph{high-norm outliers}---tokens that originate from non-informative background regions yet attract abnormally high attention, as shown in Figs.~\ref{fig:attn_artifacts} and~\ref{fig:l2_norm}. Because these high-norm outlier tokens are highly redundant in both feature and spatial dimensions, retaining them severely wastes the limited token budget and suppresses the preservation of truly informative, low-norm tokens. Second, text-guided pruning at the LLM decoder is harmed by attention dispersion and attention sink. When utilizing text-visual attention to select query-relevant visual tokens, the LLM decoder is frequently plagued by \emph{massive activations (attention sinks)}~\cite{visual_atten_sink25} and \textit{text-visual attention dispersion}~\cite{vispruner25, holov25}. Consequently, the decoder disproportionately anchors its attention onto a small set of semantically meaningless sink tokens, while its attention over the remaining tokens becomes highly dispersed. This noise prevents the model from forming a confident ranking of query-relevant regions, rendering attention-based selection highly unreliable, especially at aggressive compression ratios.

To mitigate these limitations, we propose \textbf{\method{}}, a \emph{training-free, cascading} pruning framework that operates on a coarse-to-fine principle. \textit{\method{}} fundamentally addresses the aforementioned bottlenecks by conditioning the visual stream before language decoding. 
Specifically, it first employs a \textbf{visual sanitizer} to preemptively filter high-norm outliers to yield a purified and diverse set of informative tokens. Following this purification, \textit{\method{}} applies a \textbf{text-guided pruner} at the early layers of the LLM decoder. This pre-filtering effectively reduces attention sink~\cite{visual_atten_sink25} and attention dispersion, enabling the cross-modal attention mechanism to reliably identify and retain tokens that are genuinely aligned with the textual instruction.

Comprehensive experiments demonstrate the effectiveness, efficiency, and generalizability of \textit{\method{}} across a wide range of image and video understanding tasks. Seamlessly integrated into distinct MLLM architectures---including the fixed-grid LLaVA family and dynamic-resolution Qwen2.5-VL---\textit{\method{}} yields superior accuracy-latency trade-offs compared to SOTA methods\rev{, as summarized in Fig.~\ref{fig:radar}}. Notably, it enables an 88.9\% visual token reduction while preserving 96.5\% performance on LLaVA-1.5, unlocking substantial inference speedups for resource-constrained environments. Moreover, our proposed \textit{visual sanitizer} exhibits strong transferability; it functions as an orthogonal, plug-and-play module that can be integrated to consistently enhance existing vision-centric pruning frameworks. Finally, through in-depth empirical analysis, 
% we answer a fundamental question: \textit{why does our cascading strategy outperform state-of-the-art text-guided methods?} 
we validate that \textit{\method{}} successfully reduces massive activations---cutting the ratio of attention sink tokens within the LLM decoder by about 73\%---and significantly lowers the entropy of text-visual attention, thereby empowering the downstream \textit{text-guided pruner} to operate with unprecedented confidence and precision.

\section{Related Work}
\label{sec:related}

% \subsection{MLLMs and Their Challenges}
\noindent\textbf{MLLMs and Their Challenges.}
The rapid evolution of MLLMs~\cite{DBLP:conf/nips/LiuLWL23a, Mini_Gemini24, blip_23,DBLP:conf/cvpr/LinYP0SH24,DBLP:journals/tmlr/ZhangWLLMLL25,DBLP:conf/iclr/Ye0LH0000025,DBLP:conf/iccv/HuLLW23,DBLP:conf/acl/0001RKK24,DBLP:conf/cvpr/DeitkeC0T0PSMLS25,DBLP:journals/corr/abs-2508-18265,DBLP:conf/cvpr/LiuZSZLYXCGLLTF25,DBLP:conf/nips/Dai0LTZW0FH23,zheng2026learning} has revolutionized visual understanding by bridging powerful LLMs~\cite{DBLP:conf/nips/BrownMRSKDNSSAA20,llama_tech,gpt4_tech,gpt4o,gpt5,DBLP:journals/corr/abs-2312-11805,DBLP:journals/corr/abs-2403-05530,DBLP:journals/corr/abs-2507-06261,DBLP:journals/corr/abs-2412-15115,qwen3} with advanced vision encoders~\cite{clip2021,DBLP:conf/iccv/ZhaiM0B23,DBLP:conf/cvpr/FangWXSWW0WC23,DBLP:journals/corr/abs-2504-13181,DBLP:conf/iccv/LiuL00W0LG21}. Representative architectures (\eg, LLaVA family~\cite{llava1_5, Llava_next,DBLP:journals/tmlr/0080ZGZ00ZZL0L25} and Qwen-VL series~\cite{qwen_vl23, qwen25,DBLP:journals/corr/abs-2511-21631}) have demonstrated enhanced adaptability across diverse multimodal tasks. 
To mitigate visual hallucinations and improve fine-grained perception~\cite{vlm_hallucination24,MARVEL24}, recent trends favor increasing input resolution and supporting dynamic aspect ratios. However, this strategy inevitably exacerbates the token explosion problem. For instance, while LLaVA-1.5~\cite{llava1_5} encodes a standard image into 576 tokens, high-resolution models like LLaVA-NeXT~\cite{Llava_next} scale this to over 2,880 tokens---an order of magnitude longer than typical text prompts. The situation becomes even more prohibitive for video understanding; a 1-hour video sampled at 1fps can exceed 2 million visual tokens~\cite{Video_llava}. These massive visual sequences occupy a disproportionate share of the LLM's context window, highlighting the urgent need for efficient visual token compression strategies.

% Fig. 2: Distribution & Rank
\begin{figure}[t!]
    \centering
    \begin{subfigure}[t]{0.4\linewidth}
        \centering
        \includegraphics[width=\linewidth,height=3cm,keepaspectratio]{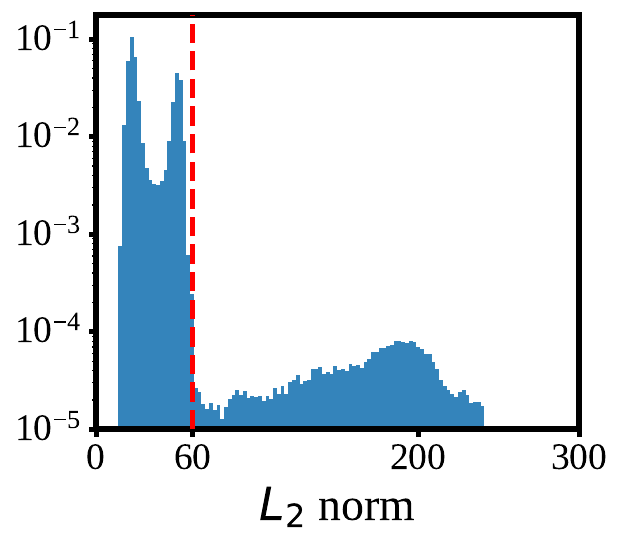}
        \caption{Feature norm distribution}
        \label{fig:feature_norm_distri}
    \end{subfigure}
    \hfill
    \begin{subfigure}[t]{0.48\linewidth}
        \centering
        \includegraphics[width=\linewidth,height=3cm,keepaspectratio]{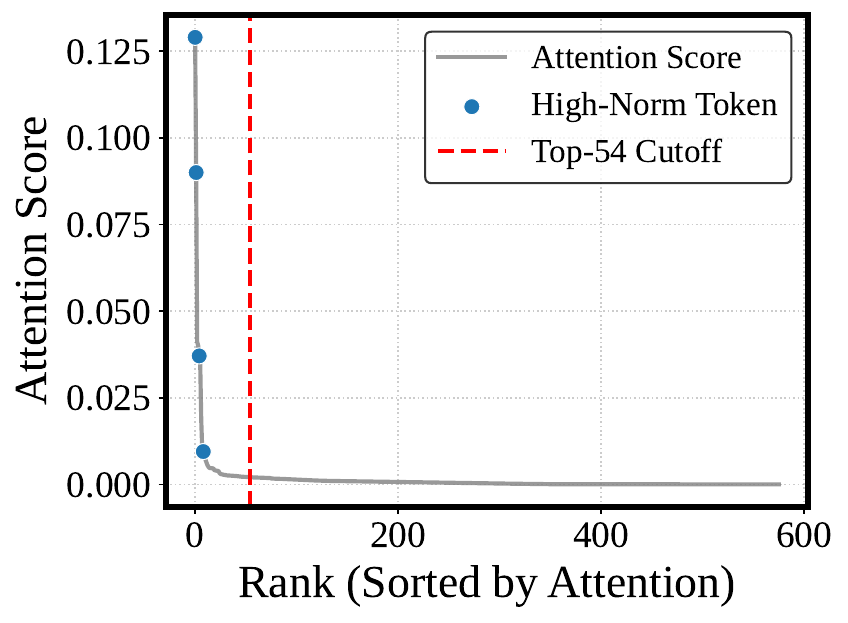}
        \caption{Attention rank of high-norm tokens}
        \label{fig:high_norm_rank}
    \end{subfigure}
    
    \caption{\textbf{Characterizing High-Norm Tokens.} (a) The norm histogram shows a clear two-regime separation: most tokens lie below the cutoff, while a small group of high-norm outliers forms a distinct high-value mode. (b) Despite being outliers, high-norm tokens consistently receive the highest attention.}
    \label{fig:combined_visuals_2}
    \vspace{-20pt}
\end{figure}

% \subsection{Visual Token Compression and Pruning}
\smallskip
\noindent\textbf{Visual Token Compression and Pruning.}
To mitigate the inefficiency caused by excessively long token sequences, extensive research has explored token compression methods across various domains, including natural language processing~\cite{DBLP:conf/icml/GoyalCRCSV20,DBLP:conf/hpca/0002Z021,DBLP:conf/naacl/YeLHS21,DBLP:conf/kdd/KimSTGKHK22} and computer vision~\cite{DBLP:conf/nips/RaoZLLZH21,DBLP:conf/cvpr/YinVAMKM22,DBLP:conf/cvpr/WeiYZTL23,DBLP:conf/ijcai/LiuWG23,DBLP:conf/iccv/0001ZLLL23,DBLP:conf/hpca/DongSLXLKMLLFW23,DBLP:conf/eccv/FayyazKJSJSPG22,DBLP:conf/cvpr/WangDJ24,DBLP:conf/emnlp/HuZHL024}. In the context of MLLMs, recent studies have primarily focused on reducing visual redundancy~\cite{DBLP:conf/aaai/LinLLJ25,DBLP:conf/cvpr/XingHDL0ZCHWWL25,DBLP:journals/corr/abs-2412-03248,DBLP:conf/cvpr/TaoQYSW25}, which can be broadly categorized into vision-centric and text-guided strategies. 
Early vision-centric approaches like ToMe~\cite{ToMe23} and PruMerge~\cite{prumerge24} employ token merging or clustering based on feature similarity to reduce sequence length without training. Building on this, importance-based pruning methods such as VisionZip~\cite{VisionZip25}, Vispruner~\cite{vispruner25}, and FasterVLM~\cite{FasterVLM24} prioritize tokens exhibiting high correlations to the [\texttt{CLS}] token. To further optimize the token dropping process, recent methods like MustDrop~\cite{MustDrop24} and PDrop~\cite{PDdrop24} introduce progressive dropping mechanisms to adaptively trim the visual sequence, while works like HoloV~\cite{holov25} utilize a holistic mechanism to retain global context during aggressive pruning. \rev{More recently, HiDrop~\cite{hidrop26} schedules \emph{when} visual tokens enter, are pruned, and exit the LLM layers via late vision injection and early vision exit, while AutoPrune~\cite{autoprune25} instead decides \emph{how many} tokens to keep per layer by mapping visual-text mutual information to a budget-constrained retention curve.}
Parallel to vision-centric methods, text-guided approaches leverage the LLM's semantic capabilities to select tokens. FastV~\cite{FastV24} and SparseVLM~\cite{SparseVLM25} focus on query-aware token selection using decoder attention scores or cross-modal guidance. 
Despite their diverse strategies, these methods often overlook the pathological high-norm outlier tokens in vision encoders\rev{~\cite{DBLP:conf/iclr/DarcetOMB24}} or the attention sink phenomenon in LLM decoders. In contrast, our \textit{\method{}} adopts a cascading framework that preemptively filters inherent visual redundancy and high-norm outliers to clarify the attention landscape, thereby enabling more precise and confident text-guided pruning.

% \section{Motivation}
% \label{sec:motivation}

% \begin{figure}[t!]
%     \centering
%     \includegraphics[width=0.99\linewidth]{figs/pipeline.pdf}
%     \caption{\textbf{Overview of \textit{\method{}} framework}. 
%     \textbf{Left (Visual Sanitizer):} As shown in the \textit{Attention Ranking} (far left), \textit{high-norm outlier tokens} (purple) dominate the top ranks despite being background noise. We aggregate these outliers into a single \textit{sink token} (red hashed) and select informative \textit{reserved tokens} (orange) via attention and similarity filtering.
%     \textbf{Right (Text-Guided Pruner):} The purified visual sequence further interacts with \textit{text tokens} (green), utilizing accumulated text-to-vision attention to retain only semantically relevant visual tokens.}
%     \label{fig:pipeline}
%     \vspace{-15pt}
% \end{figure}

\begin{figure*}[t!]
    \centering
    \includegraphics[width=0.95\textwidth]{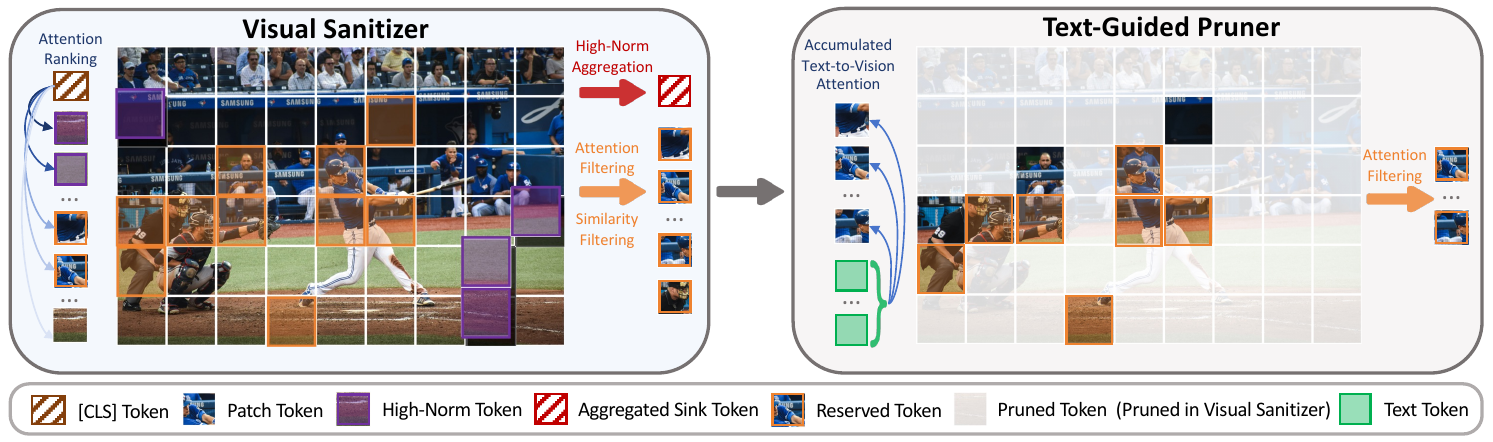}
    \caption{\textbf{Overview of \textit{\method{}} framework}. 
    \textbf{Left (Visual Sanitizer):} As shown in the \textit{Attention Ranking} (far left), \textit{high-norm outlier tokens} (purple) dominate the top ranks despite being background noise. We aggregate these outliers into a single \textit{sink token} (red hashed) and select informative \textit{reserved tokens} (orange) via attention and similarity filtering.
    \textbf{Right (Text-Guided Pruner):} The purified visual sequence further interacts with \textit{text tokens} (green), utilizing accumulated text-to-vision attention to retain only semantically relevant visual tokens.}
    \label{fig:pipeline}
    \vspace{-10pt}
\end{figure*}

\section{Methodology}
\label{sec:method}
In this section, we detail the motivation and architecture of \textbf{\method{}}, a novel, training-free visual token pruning framework for efficient MLLM inference. 
First, we analyze efficiency bottlenecks of MLLMs, highlighting the critical need to compress extensive visual token sequences. 
We then present key empirical observations revealing that \textbf{high-norm outlier tokens} are inherently redundant and act as attention sinks that disrupt vision-centric pruning.
Motivated by these insights, we introduce our coarse-to-fine cascading \textit{\method{}}, as shown in Fig.~\ref{fig:pipeline}, which features a \textbf{visual sanitizer} and a \textbf{text-guided pruner}.
% which decouples token reduction into two synergistic stages. 
% Specifically, a \textbf{visual sanitizer} first preemptively filters these high-norm outliers to yield a purified and diverse set of informative tokens. Second, leveraging this clean representation, we use \textbf{text-guided pruning} in the early LLM layers evaluates cross-modal alignment to precisely retain the tokens most essential for textual reasoning.
Specifically, the \textit{visual sanitizer} first preemptively filters high-norm outliers to yield a purified and diverse set of informative tokens, and the \textit{text-guided pruner} subsequently retains visual tokens that are semantically aligned with the text query.

%-------------------------------------------------------------------------
% Overview Paragraph
%-------------------------------------------------------------------------

\subsection{Preliminary}
\label{sec:pre}

\noindent \textbf{Architecture of MLLMs.} Existing MLLMs generally consist of a vision encoder and an LLM, both built on the Transformer architecture~\cite{vaswani2017attention}. 
To be specific, the vision encoder (\eg, CLIP-ViT~\cite{clip2021}) applies self-attention over visual tokens $\mX_{vis} =[{\vx}_{\text{CLS}}; {\vx}_{\text{img}}^{1}, \dots, {\vx}_{\text{img}}^{n_{vis}-1}]$, where $n_{vis}$ denotes the total number of visual tokens. The attention matrix is computed as $\mA_{vis} = \text{softmax}(\mQ {\mK}^{T} / \sqrt{d})$, where $\mQ$ and $\mK$ are the query and key representations, and $d$ represents the hidden dimension. We denote the first row of $\mA_{vis}$---representing the attention from the [\texttt{CLS}] token to all image patches---as \textit{\texttt{CLS} attention}.
Conversely, the LLM processes the concatenated input sequence $\mX =[{\vx}_{\text{sys}}; \mX_{vis}; {\vx}_{\text{txt}}]$ using causal self-attention. Here, ${\vx}_{\text{sys}}$ and ${\vx}_{\text{txt}}$ denote the system prompt and text instruction with sequence lengths $n_{sys}$ and $n_{text}$, respectively. This brings the total sequence length to $n = n_{sys} + n_{vis} + n_{text}$. This causal mechanism incorporates rotary position embeddings (RoPE) $\mR_{\theta}$ parameterized by $\theta$, along with a lower-triangular causal mask $\mM$ to ensure unidirectional routing. Accordingly, the LLM attention matrix is computed as $\mA_{llm} = \text{softmax}\big(({\mR}_{\theta}\mQ) {({\mR}_{\theta}{\mK})}^{T} / \sqrt{d} + \mM\big)$, and the final output is obtained via $\mO = \mA_{llm} \mV$.

% \begin{figure}[t!]
%     \centering
%     % --- 左图：约 1/4 宽度 ---
%     \begin{subfigure}[b]{0.36\linewidth}
%         \centering
%         % 你的密度分布图
%         \includegraphics[width=\linewidth]{figs/neighbor_sim_distribution.pdf}
%         % \label{fig:neighbor_sim} % 如果需要单独引用左图可取消注释
%     \end{subfigure}
%     \hfill % 自动填充左右间距
%     % --- 右图：约 3/4 宽度 ---
%     \begin{subfigure}[b]{0.62\linewidth}
%         \centering
%         % 你的特征相似度矩阵/图
%         \includegraphics[width=\linewidth]{figs/Sample_8_Similarity.pdf}
%         %\label{fig:feature_sim} % 如果需要单独引用右图可取消注释
%     \end{subfigure}

%     % --- 统一的主标题 ---
%     \caption{\textbf{Feature Redundancy Analysis.} Left: high-norm tokens show distinctively high neighbor similarity. Right: High-norm tokens exhibit extremely high inter-token similarity (feature space collapse), while low-norm tokens maintain high diversity.}
%     \label{fig:high_norm_analysis}
%     \vspace{-18pt}
% \end{figure}

\begin{figure}[t!]
    \centering

    % --- Top ---
    \begin{subfigure}[t]{0.6\linewidth}
        \centering
        \includegraphics[width=\linewidth]{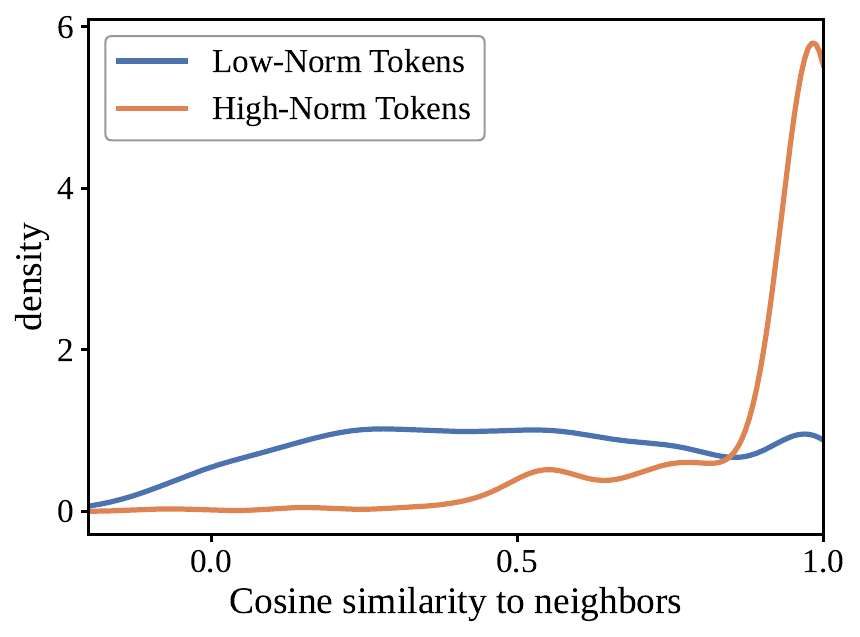}
        \caption{Neighbor similarity}
        \label{fig:neighbor_sim}
    \end{subfigure}

    \vspace{0.4em}

    % --- Bottom ---
    \begin{subfigure}[t]{0.9\linewidth}
        \centering
        \includegraphics[width=\linewidth]{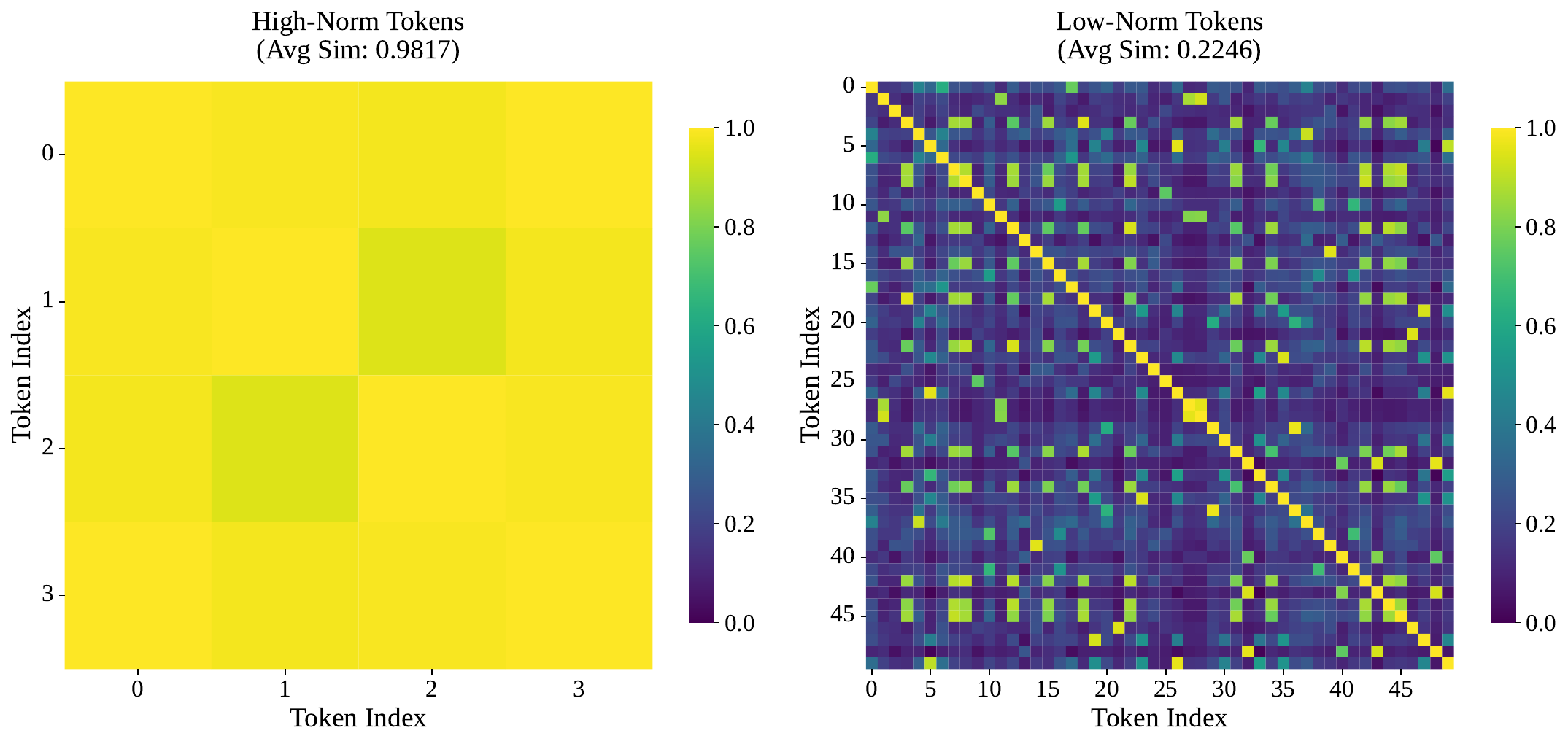}
        \caption{Pairwise similarity within each subset}
        \label{fig:pairwise_sim}
    \end{subfigure}

    \caption{\textbf{Feature Redundancy Analysis.}
    \rev{(a)} high-norm tokens show distinctively high neighbor similarity.
    \rev{(b)} high-norm tokens exhibit extremely high inter-token similarity (feature space collapse), while low-norm tokens maintain high diversity.}
    \label{fig:high_norm_analysis}
    \vspace{-20pt}
\end{figure}

% \subsubsection{Computation Complexity.}
\smallskip
\noindent \textbf{Efficiency Bottleneck in MLLMs.} 
Computing the aforementioned attention matrix $\mA_{llm}$ and processing the concatenated sequence $\mX$ through deep LLM layers introduces severe computational overhead. Specifically, for an LLM with $T$ layers and a feed-forward network intermediate width $m$, the total floating-point operations (FLOPs) can be approximated as $T \times (4nd^2 + 2n^2 d + 2ndm)$. This formulation highlights a quadratic computational complexity with respect to the total sequence length $n$, primarily stemming from the causal self-attention mechanism (\ie, the $2n^2d$ term). Given that $n = n_{sys} + n_{vis} + n_{text}$, this sequence length in typical MLLM applications is dominated by the visual tokens $n_{vis}$, which often exceed text prompts by an order of magnitude. Because this massive volume of visual tokens acts as the primary catalyst for the quadratic computational explosion, \textbf{reducing the visual token count $n_{vis}$} emerges as the most essential paradigm for accelerating MLLM inference.

% \subsubsection{The ``Attention Sink'' Trap.} 
% A critical observation is the behavior of these tokens within the self-attention mechanism. As shown in Fig.~\ref{fig:high_norm_rank}, high-norm tokens consistently receive the highest attention scores from the [CLS] token. This phenomenon suggests they function as "attention sinks," absorbing a disproportionate amount of global attention.
% This behavior poses a fundamental challenge for existing purely-visual pruning methods (\eg, VisionZip~\cite{VisionZip25}, FasterVLM~\cite{FasterVLM24}). Since these methods rely exclusively on attention scores to determine token importance, they retain them with the highest priority while discarding potentially more informative low-norm tokens.

% \subsection{Information Redundancy in high-norm outlier tokens}
% \subsection{Information Redundancy in High-Norm Outlier Tokens}
\subsection{Redundancy in High-Norm Tokens}
\label{sec:motivation}

As established in Sec.~\ref{sec:pre}, reducing visual tokens is essential for efficient MLLM inference. To achieve this, existing pruning methods predominantly rely on attention scores (\eg, \textit{\texttt{CLS} attention}) to gauge token importance, implicitly assuming that tokens with higher attention weights carry richer semantic information. However, the reliability of this assumption remains questionable. Upon inspecting the attention maps within the vision encoder of MLLMs, we observe a counterintuitive phenomenon: clear outliers emerge, where a few non-semantic background regions exhibit unusually peaky attention values. Since attention-based pruning methods are highly sensitive to these peaky values, understanding the true nature of these outliers---and whether they actually encode indispensable visual concepts---is crucial for designing a robust compression strategy.

% \begin{figure}[t!]
% \centering
%     \begin{subfigure}[t]{0.48\linewidth}
%         \centering
%         \includegraphics[width=\linewidth]{figs/Sample_11_Spatial.pdf}
%         \caption{}
%         \label{fig:high_norm_place}
%     \end{subfigure}
%     \hfill
%     \begin{subfigure}[t]{0.48\linewidth}
%         \centering
%         \includegraphics[width=\linewidth]{figs/Sample_199_Spatial.pdf}
%         \caption{}
%         \label{fig:high_norm_place1}
%     \end{subfigure}
%     \caption{\textbf{Spatial Redundancy Analysis.}
%     Visualizations demonstrate that high-norm outlier tokens (red) predominantly appear in non-semantic regions with high local similarity, whereas filtered informative low-norm tokens (blue, top-ranked by \textit{\textit{\texttt{CLS} attention}}) correspond to unique local features.}
%     \label{fig:spatial_redundancy}
%     \vspace{-18pt}
% \end{figure}

% \subsubsection{Outliers are high-norm tokens.}
\smallskip
\noindent\textbf{Outliers are high-norm tokens.}
We first seek a quantitative way to localize these outliers. By inspecting the vision encoder features, we observe that artifact regions are strongly associated with unusually large feature magnitudes\rev{, an effect first reported for self-supervised and supervised ViTs by \citet{DBLP:conf/iclr/DarcetOMB24}, who mitigate it at training time by appending dedicated register tokens. We instead exploit it at inference time as a training-free signal for token pruning}. As shown in Figs.~\ref{fig:attn_artifacts} and~\ref{fig:l2_norm}, tokens from artifact patches exhibit much larger $\ell_2$ norms than tokens from semantic regions.
To further verify this observation, we plot the distribution of token norms over a subset of images in Fig.~\ref{fig:feature_norm_distri}. The histogram shows a clear two-regime pattern: the majority of tokens concentrate below $60$, while a separated group of outliers spans a much higher range (roughly $[60, 250]$), \rev{which accounts for only about $1\%$ of all visual tokens. Since the absolute norm scale is encoder-dependent, we identify outliers by their relative rank rather than by a hand-tuned cutoff: given the $N$ visual tokens of an image, we rank them by feature norm and treat the top $\rho$ fraction as \textbf{high-norm} tokens, \ie, $\mathbf{X}_{high}=\mathrm{Top}_{\rho}\big(\{\|x_i\|_2\}_{i=1}^{N}\big)$ with $|\mathbf{X}_{high}|=\lceil \rho N \rceil$, and the remaining tokens $\mathbf{X}_{low}=\mathbf{X}\setminus\mathbf{X}_{high}$ as low-norm tokens. Following the observed ratio above, we set $\rho=1\%$ throughout, which keeps the criterion applicable to encoders whose norms live on very different scales (see Appendix~\ref{app:nonclip_norm}).} Importantly, these high-norm tokens largely overlap with the observed outliers.

\smallskip
\noindent\textbf{High-norm outlier tokens are spatially redundant.}
Next, we examine whether high-norm outlier tokens contain informative visual content. We measure the cosine similarity between each token and its 4-nearest spatial neighbors after the patch embedding layer. As shown in Fig.~\ref{fig:neighbor_sim}, high-norm outliers typically arise from patches that are highly similar to their neighboring patches. This observation aligns with the visual pattern that these outliers predominantly appear in homogeneous background regions (\eg, sky and walls)\rev{; see Appendix~\ref{app:more_visuals} for visualization results}. Therefore, these tokens encode spatially redundant information and can be discarded without degrading image representation.

% \subsubsection{High-norm outlier tokens are feature-redundant.}
\smallskip
\noindent\textbf{High-norm outlier tokens are feature-redundant.}
We further study redundancy in feature space. Standard attention-based pruning keeps the top-$k$ tokens ranked by \textit{\textit{\texttt{CLS} attention}}, but this set is heavily dominated by high-norm tokens. We therefore partition the selected tokens into the high-norm subset $\mathbf{X}_{high}$ and the low-norm subset $\mathbf{X}_{low}$ \rev{defined above}.
We then compute pairwise cosine similarity within each subset (Fig.~\ref{fig:pairwise_sim}). High-norm tokens show extremely high intra-set similarity, indicating representational collapse. In contrast, low-norm tokens remain more diverse and typically carry richer semantic details; see the Appendix~\ref{app:more_visuals} for visualizations. This confirms that retaining high-norm tokens wastes the token budget on redundant information.

% \subsubsection{The ``attention sink'' trap.}
\smallskip
\noindent\textbf{The ``attention sink'' trap.}
Although high-norm outlier tokens are largely redundant, they can still dominate self-attention. As shown in Fig.~\ref{fig:high_norm_rank}, they consistently receive the highest attention from the [\texttt{CLS}] token, acting as \emph{attention sinks} that absorb a disproportionate amount of global attention. This creates a failure mode for vision-centric pruning methods such as VisionZip~\cite{VisionZip25} and FasterVLM~\cite{FasterVLM24}: when token importance is determined solely by attention scores, these methods prioritize high-norm outlier tokens while potentially discarding more informative low-norm tokens.

\subsection{\method{} Framework}

Building upon observations in Sec.~\ref{sec:motivation}, we propose \textbf{\method{}}, a training-free, cascading visual token pruning framework designed for efficient MLLM inference, as shown in Fig.~\ref{fig:pipeline}. Our approach operates on a coarse-to-fine principle: we first introduce a \textbf{visual sanitizer} to filter high-norm outliers, yielding a purified and diverse set of informative tokens, and then a \textbf{text-guided pruner} to retain visual tokens that are semantically aligned with the text query.

\smallskip
\noindent\textbf{Visual sanitizer.}
To address the ``high-norm trap'' where high-norm outlier tokens act as attention sinks, we explicitly isolate these tokens based on feature magnitude. Let $\mathbf{X} = \{x_1, \dots, x_N\} \in \mathbb{R}^{N \times D}$ denote a sequence of visual tokens extracted from the vision encoder. 
We first compute the $L_2$-norm $n_i=\|x_i\|_2$ for each visual token, and \rev{apply the scale-free top-$\rho$ rule of Sec.~\ref{sec:motivation} to partition the visual sequence into outliers $\mathbf{X}_{high}=\mathrm{Top}_{\rho}(\{n_i\}_{i=1}^{N})$, \ie, the $\lceil \rho N \rceil$ tokens with the largest norms, and candidate tokens $\mathbf{X}_{low}=\mathbf{X}\setminus\mathbf{X}_{high}$, with $\rho=1\%$ by default. Since this rule ranks tokens against each other instead of against an absolute value, it requires no per-model calibration and transfers directly to backbones whose feature norms live on different scales.}

Subsequently, we aggregate the identified high-norm outlier tokens $\mathbf{X}_{high}$ into a single proxy token to compress visual redundancy, while preserving global information. 
To be specific, \rev{as discussed in Sec.~\ref{sec:motivation},} these tokens are compressed without significant information loss via average pooling due to their extreme feature similarity:
\begin{equation}
    x_{sink} = \frac{1}{|\mathbf{X}_{high}|} \sum_{x \in \mathbf{X}_{high}} x.
\end{equation}
% \vspace{2pt}

% \noindent\textbf{Salience-Diversity Selection for Low-Norm Tokens.}
We then employ a hybrid selection strategy to extract a representative subset from the informative candidates $\mathbf{X}_{low}$, balancing semantic salience with spatial coverage. 
First, to preserve prominent regions, we select the set $\mathbf{X}_{res}$ containing the top-$k_{res}$ tokens with the highest \textit{CLS attention} scores $\mathcal{A}_{cls}$\rev{\footnote{\rev{For encoders without a [\texttt{CLS}] token (\eg, the vision tower of Qwen2.5-VL~\cite{qwen25}), $\mathcal{A}_{cls}$ is replaced by the average received visual self-attention, while all other components remain unchanged; see Appendix~\ref{app:cls_free}.}}}:
\begin{equation}
    \mathbf{X}_{res} = \left\{ x \in \mathbf{X}_{low} \mid \text{Rank}(\mathcal{A}_{cls}(x)) \le k_{res} \right\}.
\end{equation}
Next, to capture subtle background semantics and avoid feature collapse, we perform a similarity-based de-duplication on the remaining tokens $\mathcal{R} = \mathbf{X}_{low} \setminus \mathbf{X}_{res}$. We initialize the selected set $\mathcal{S} = \mathbf{X}_{res}$ and iteratively\footnote{\rev{In practice we use a batched approximation of this serial rule; see Appendix~\ref{app:batched_selection}.}} sample a token $x^*$ that exhibits the lowest similarity to the current set $\mathcal{S}$, appending it to $\mathbf{X}_{div}$:
\begin{equation}
    x^* = \operatorname*{argmin}_{x \in \mathcal{R}} \left( \max_{s \in \mathcal{S}} \text{CosSim}(x, s) \right).
    \label{eq:diversity}
\end{equation}

The final purified visual representation $\mathbf{Z}$ is formed by concatenating the aggregated sink token and the selected subsets: $\mathbf{Z} = [x_{sink}, \mathbf{X}_{res}, \mathbf{X}_{div}]$, which is subsequently projected and fed into the LLM.

% \subsubsection{Text-guided pruner.}
\smallskip
\noindent\textbf{Text-guided pruner.}
Leveraging the purified sequence $\mathbf{Z}$, the model effectively evaluates semantic relevance without interference from high-norm outliers. We utilize the text-to-vision attention in an early LLM decoder layer as the importance estimator. Letting $L_t$ denote the number of text tokens, we compute a global relevance score $\tilde{p}_j$ for each visual token $z_j \in \mathbf{Z}$ by aggregating its attention weights across all textual queries:
\begin{equation}
    \tilde{p}_j = \frac{1}{L_t} \sum_{i=1}^{L_t} \text{Softmax}(\mathbf{Q}_{text} \cdot \mathbf{K}_{vis}^\top)_{i,j}.
\end{equation}
% where $L_t$ denotes the number of text tokens.
We then retain the top-$K$ visual tokens with the highest $\tilde{p}$ scores. This query-aware selection ensures the LLM focuses its computational budget strictly on regions required for reasoning.

\section{Experiments}
\label{sec:exp}
\renewcommand{\multirowsetup}{\centering}
\definecolor{mygray}{gray}{.92}
\definecolor{mygreen1}{RGB}{253, 244, 244}
\definecolor{mygreen2}{RGB}{238, 243, 243}
\definecolor{ForestGreen}{RGB}{34,139,34}
\newcommand{\fg}[1]{\mathbf{\mathcolor{ForestGreen}{#1}}}
\definecolor{Forestred}{RGB}{220,50,50}
\newcommand{\fr}[1]{\mathbf{\mathcolor{Forestred}{#1}}}

\begin{table}[t]
    \centering
    \setlength{\tabcolsep}{3.5pt}
    \scriptsize
    
    \resizebox{\linewidth}{!}{
    \begin{tabular}{l | *{7}{>{\centering\arraybackslash}p{1.1cm}} |>{\centering\arraybackslash}p{1.2cm}}
    \toprule
        \textbf{\;Methods} & \textbf{GQA} & \textbf{MMB} & \textbf{MMB}$_{\text{CN}}$ & \textbf{MME} & \textbf{POPE} & \textbf{SQA} & \textbf{VQA}$_{\text{Text}}$ & \makecell[c]{\textbf{Avg}}\\
        \midrule
        
        % Upper bound
        \textcolor{gray}{Upper Bound, 576 Tokens} 
        & \textcolor{gray}{61.9} 
        & \textcolor{gray}{64.7} 
        & \textcolor{gray}{58.1} 
        & \textcolor{gray}{1862} 
        & \textcolor{gray}{85.9} 
        & \textcolor{gray}{69.5} 
        & \textcolor{gray}{58.2} 
        & \textcolor{gray}{100.0\%} \\
        \midrule

        % ================= Retain 64 ===================
        \rowcolor{mygray}
        LLaVA-1.5-7B & \multicolumn{8}{c}{\textit{Retain 64 Tokens} \ $\fg{(\text{Pruning Ratio}= 88.9\%)}$}\\

        ToMe \texttt{\scriptsize{(ICLR23)}} 
        & 48.6 & 43.7 & - & 1138 & 52.5 & 50.0 & 45.3 & 59.7 \\

        FastV \texttt{\scriptsize{(ECCV24)}} 
        & 46.1 & 48.0 & 52.7 & 1256 & 48.0 & 51.1 & 47.8 & 74.1 \\

        MustDrop \texttt{\scriptsize{(2024.11)}} 
        & 53.1 & 60.0 & 53.1 & 1612 & 68.0 & 63.4 & 54.2 & 88.6 \\

        PDrop \texttt{\scriptsize{(2024.10)}} 
        & 41.9 & 33.3 & 50.5 & 1092 & 55.9 & 68.6 & 45.9 & 72.5 \\

        VisionZip \texttt{\scriptsize{(CVPR25)}} 
        & 55.1 & 60.1 & 55.4 & 1690 & 77.0 & 69.0 & \textbf{55.5} & 93.2 \\

        SparseVLM \texttt{\scriptsize{(ICML25)}} 
        & 52.7 & 56.2 & 46.1 & 1505 & 75.1 & 62.2 & 51.8 & 85.4 \\

        HoloV \texttt{\scriptsize{(NeurIPS25)}} 
        & 55.3 & \textbf{63.3} & 55.1 & 1715 & 80.3 & 69.5 & 55.4 & 94.7 \\

        ApET \texttt{\scriptsize{(2026.02)}} 
        & 56.9 & 61.2 & 54.4 & 1714 & \textbf{84.4} & 68.9 & 53.9 & 94.6 \\

        \rowcolor{mygreen2}
        \textbf{\method{} (Ours)} 
        & \textbf{57.4} & 62.8 & \textbf{56.9} & \textbf{1754} & 83.8 & \textbf{70.0} & \textbf{55.5} & \textbf{96.5} \\
        
        \midrule

        % ================= Retain 32 ===================
        \rowcolor{mygray}
        LLaVA-1.5-7B & \multicolumn{8}{c}{\textit{Retain 32 Tokens} \ $\fg{(\text{Pruning Ratio}= 94.4\%)}$}\\

        ToMe \texttt{\scriptsize{(ICLR23)}} 
        & 43.6 & 31.6 & 28.1 & 828 & 39.0 & 41.4 & 38.3 & 54.7 \\

        FastV \texttt{\scriptsize{(ECCV24)}} 
        & 41.5 & 37.8 & 33.2 & 885 & 32.5 & 42.6 & 42.5 & 57.5 \\

        SparseVLM \texttt{\scriptsize{(ICML25)}} 
        & 48.3 & 51.4 & 40.6 & 1047 & 67.9 & 57.3 & 46.1 & 74.9 \\

        PruMerge+ \texttt{\scriptsize{(2024.05)}} 
        & 51.1 & 56.8 & 47.0 & 941 & 70.9 & 68.5 & 50.6 & 81.4 \\

        VisionZip \texttt{\scriptsize{(CVPR25)}} 
        & 51.8 & 57.7 & 50.3 & 1247 & 68.7 & 68.8 & 53.1 & 85.2 \\

        VisPruner \texttt{\scriptsize{(ICCV25)}} 
        & 52.2 & 58.4 & 52.7 & 1271 & 72.7 & 69.2 & 53.9 & 87.2 \\

        \rowcolor{mygreen2}
        \textbf{\method{} (Ours)} 
        & \textbf{55.1} & \textbf{61.2} & \textbf{55.3} & \textbf{1363} & \textbf{78.5} & \textbf{69.9} & \textbf{54.8} & \textbf{91.2} \\
        \bottomrule
    \end{tabular}
    }
    \caption{Performance comparison on LLaVA-1.5-7B across different image-language benchmarks under different pruning ratios. Best results are in \textbf{bold}. More comprehensive results are provided in the Appendix~\ref{app:additional_exps}.}
    \label{tab:llava1-5}
    \vspace{-10pt}
\end{table}
\renewcommand{\multirowsetup}{\centering}
\definecolor{mygray}{gray}{.92}
\definecolor{mygreen2}{RGB}{238, 243, 243}
\definecolor{ForestGreen}{RGB}{34,139,34}
\renewcommand{\fg}[1]{\mathbf{\mathcolor{ForestGreen}{#1}}}

\begin{table}[t]
    \centering
    \setlength{\tabcolsep}{2.0pt}
    \tiny

    \begin{tabular}{l | *{5}{>{\centering\arraybackslash}p{0.85cm}} |>{\centering\arraybackslash}p{0.5cm}}
    \toprule
        \textbf{Methods} 
        & \textbf{MMB} & \textbf{MME} & \textbf{POPE} & \textbf{SQA} & \textbf{VQA$_{\text{Text}}$} & {\textbf{Avg}}\\
        \midrule
        
        \textcolor{gray}{Upper Bound} 
        & \textcolor{gray}{84.2} 
        & \textcolor{gray}{2292} 
        & \textcolor{gray}{86.1} 
        & \textcolor{gray}{88.78} 
        & \textcolor{gray}{83.04}  
        & \textcolor{gray}{100\%} \\
        \midrule

        % 66.7%
        \rowcolor{mygray}
        Qwen2.5-VL-7B & \multicolumn{6}{c}{\textit{Pruning Ratio = $\fg{\downarrow 66.7\%}$}}\\

        FastV \texttt{\scriptsize{(ECCV24)}} 
        & 75.7 & 2072 & 82.2 & 78.5 & 77.9 & 91.6 \\

        HoloV \texttt{\scriptsize{(NeurIPS25)}} 
        & 78.3 & 2093 & 85.0 & 79.8 & 78.9 & 93.6 \\

        VisionZip \texttt{\scriptsize{(CVPR25)}}
        & 75.8 & 2098 & 84.4 & 81.7 & 70.9 & 91.4 \\

        \rowcolor{mygreen2}
        \textbf{\method{} (Ours)} 
        & \textbf{82.1} & \textbf{2268} & \textbf{85.2} & \textbf{88.7} & \textbf{81.1} & \textbf{98.6} \\
        \midrule
        
        % 77.8%
        \rowcolor{mygray}
        Qwen2.5-VL-7B & \multicolumn{6}{c}{\textit{Pruning Ratio = $\fg{\downarrow 77.8\%}$}}\\

        FastV \texttt{\scriptsize{(ECCV24)}} 
        & 74.9 & 2036 & 80.7 & 78.0 & 69.0 & 88.5 \\

        HoloV \texttt{\scriptsize{(NeurIPS25)}} 
        & 76.5 & 2043 & 82.3 & 79.8 & 70.3 & 90.0 \\

        % TODO-CHECK: the VisionZip row below was UPDATED to the re-run numbers
        % reported in the rebuttal (Reviewer rBoA, Q2). Previous submission values were:
        % & 73.8 & 2000 & 82.6 & 79.7 & 64.8 & 87.7 \\
        % Please confirm which run should be kept, and re-run VisionZip at 66.7%
        % accordingly so that the three blocks stay monotone.
        VisionZip \texttt{\scriptsize{(CVPR25)}}
        & \textbf{80.33} & 2174 & \textbf{83.38} & 84.23 & 70.43 & 93.4 \\

        \rev{DivPrune} \texttt{\scriptsize{(CVPR25)}}
        & 76.98 & 2163 & 80.59 & 80.91 & 65.86 & 90.0 \\

        \rev{MMTok} \texttt{\scriptsize{(2025.08)}}
        & 79.30 & \textbf{2217} & 82.38 & 81.61 & 70.49 & 92.7 \\

        \rowcolor{mygreen2}
        \textbf{\method{} (Ours)} 
        & 79.7 & 2207 & 82.8 & \textbf{87.7} & \textbf{79.2} & \textbf{96.3} \\
        \midrule

        % 88.9%
        \rowcolor{mygray}
        Qwen2.5-VL-7B & \multicolumn{6}{c}{\textit{Pruning Ratio = $\fg{\downarrow 88.9\%}$}}\\

        FastV \texttt{\scriptsize{(ECCV24)}} 
        & 69.2 & 1940 & 78.6 & 77.4 & 60.3 & 83.6 \\

        HoloV \texttt{\scriptsize{(NeurIPS25)}} 
        & 72.4 & 2006 & \textbf{80.7} & 79.5 & 61.8 & 86.2 \\

        % TODO-CHECK: previous submission values for this row were:
        % & 68.0 & 1992 & 77.8 & 78.6 & 51.6 & 81.7 \\
        VisionZip \texttt{\scriptsize{(CVPR25)}}
        & 75.60 & 2003 & 78.90 & 82.30 & 63.78 & 87.7 \\

        \rev{DivPrune} \texttt{\scriptsize{(CVPR25)}}
        & 72.85 & 1957 & 74.99 & 79.57 & 59.59 & 84.1 \\

        \rev{MMTok} \texttt{\scriptsize{(2025.08)}}
        & 74.74 & 2051 & 78.75 & 80.47 & 63.90 & 87.5 \\

        \rowcolor{mygreen2}
        \textbf{\method{} (Ours)} 
        & \textbf{77.6} & \textbf{2123} & 78.3 & \textbf{87.1} & \textbf{70.8} & \textbf{91.8} \\
        
        \bottomrule
    \end{tabular}
    \caption{Performance comparison on Qwen2.5-VL-7B with various token compression methods under different pruning ratios. Best results are in \textbf{bold}. More comprehensive results are provided in the Appendix.}
    \label{tab:qwen_main}
    \vspace{-10pt}
\end{table}
\definecolor{mygreen2}{RGB}{238, 243, 243}
\begin{table}[t]
    \centering
    \scriptsize
    \setlength{\tabcolsep}{2.2pt}

    \begin{tabular}{l|ccc|c}
        \toprule
        \textbf{Method} 
        & \makecell{\textbf{MVBench}\\Acc}
        & \makecell{\textbf{SEEDBench}\\Acc}
        & \makecell{\textbf{VideoMME}\\Score}
        & \makecell{\textbf{Avg.}\\(\%)} \\
        \midrule

        \rowcolor{gray!10}
        Qwen2.5-VL-7B (Full) & 68.10 & 62.18 & 60.67 & 100.0\% \\

        DART (EMNLP25)    & 65.80 & 61.00 & 57.74 & 96.6\% \\
        DivPrune (CVPR25) & 65.85 & 59.79 & 57.78 & 96.0\% \\

        \rowcolor{mygreen2}
        \textbf{\method{} (Ours)}       & \textbf{66.70} & \textbf{61.80} & \textbf{58.59} & \textbf{98.0\%} \\

        \bottomrule
    \end{tabular}
    \caption{Performance comparison of various methods across different video-language benchmarks under an 80\% token pruning ratio. More comprehensive experimental results are provided in the Appendix~\ref{app:additional_exps}.}
    \label{tab:video_qwen25} % 确保 Label 与正文引用一致
    \vspace{-10pt}
\end{table}

% NEW (camera-ready): harder reasoning-oriented benchmarks.
% =============================================================================
% NEW TABLE for the camera-ready version.
% Source: rebuttal to Reviewer gBV3, Q7 (Evaluation on harder reasoning and
%         complex understanding tasks). LLaVA-1.5-7B, Retain-32.
% Format matches table/table_Qwen.tex (single column, \tiny).
% =============================================================================
\begin{table}[t]
    \centering
    \setlength{\tabcolsep}{2.0pt}
    \tiny

    \begin{tabular}{l | *{4}{>{\centering\arraybackslash}p{0.85cm}} |>{\centering\arraybackslash}p{0.5cm}}
    \toprule
        \textbf{Methods}
        & \textbf{MMStar} & \textbf{MMMU} & \textbf{AI2D} & \textbf{MM-Vet} & {\textbf{Avg}}\\
        \midrule

        \textcolor{gray}{Upper Bound}
        & \textcolor{gray}{34.04}
        & \textcolor{gray}{36.56}
        & \textcolor{gray}{55.15}
        & \textcolor{gray}{33.99}
        & \textcolor{gray}{100\%} \\
        \midrule

        \rowcolor{mygray}
        LLaVA-1.5-7B & \multicolumn{5}{c}{\textit{Retain 32 Tokens} \ $\fg{(\text{Pruning Ratio}= 94.4\%)}$}\\

        FastV \texttt{\scriptsize{(ECCV24)}}
        & 30.41 & 35.11 & 50.49 & 22.66 & 85.9 \\

        VisionZip \texttt{\scriptsize{(CVPR25)}}
        & 30.36 & \textbf{35.67} & 51.81 & 25.23 & 88.7 \\

        \rowcolor{mygreen2}
        \textbf{\method{} (Ours)}
        & \textbf{32.14} & 35.44 & \textbf{53.63} & \textbf{29.36} & \textbf{93.7} \\
        \bottomrule
    \end{tabular}
    \caption{\rev{Performance comparison on LLaVA-1.5-7B across harder reasoning-oriented benchmarks under an aggressive 32-token budget. Best results are in \textbf{bold}.}}
    \label{tab:harder_bench}
    \vspace{-10pt}
\end{table}

\subsection{Experiment settings}
\noindent \textbf{Evaluation benchmarks.}
To comprehensively demonstrate the effectiveness and generalizability of our \textit{\method{}}, we conduct experiments across a wide range of multimodal understanding tasks. For image-language understanding, we evaluate on \rev{twelve} widely used benchmarks\rev{: GQA~\cite{GQA19}, MMBench (MMB) and its Chinese counterpart MMB-CN~\cite{mmbench24}, MME~\cite{mme23}, POPE~\cite{pope23}, ScienceQA (SQA)~\cite{scienceqa}, VQA-v2~\cite{balanced_vqa_v2}, and TextVQA~\cite{textvqa19}, which mainly probe perception and short-form reasoning, together with four substantially harder benchmarks that target high-level reasoning rather than perception alone, namely MMStar~\cite{mmstar24}, MMMU~\cite{mmmu24}, AI2D~\cite{ai2d16}, and MM-Vet~\cite{mmvet24}}. Furthermore, since video processing inherently involves massive spatiotemporal redundancy, making efficient pruning critical, we also extend our evaluation to video-language understanding. Specifically, we conduct experiments on \rev{four} representative video benchmarks: MVBench~\cite{mvbench}, SEED-Bench~\cite{seed_bench23}, \rev{NextQA~\cite{nextqa21},} and VideoMME~\cite{videomme}. \rev{We describe each benchmark and the corresponding evaluation metric in Appendix~\ref{app:benchmarks_metrics}.} 

% \smallskip
% \noindent \textbf{Comparison methods.} We compare \textit{\method{}} with existing state-of-the-art (SoTA) visual token reduction methods based on different MLLM architectures (\ie, LLaVA-1.5~\cite{llava1_5} and Qwen2.5-VL~\cite{qwen25}). These baselines employ diverse strategies---such as token merging, attention-based pruning, adaptive allocation, and hierarchical retention---to improve efficiency by reducing redundant tokens. Specifically, we compare against ToMe~\cite{ToMe23}, MustDrop~\cite{MustDrop24}, FastV~\cite{FastV24}, PDrop~\cite{PDdrop24}, PruMerge~\cite{prumerge24}, SparseVLM~\cite{SparseVLM25}, VisionZip~\cite{VisionZip25}, Vispruner~\cite{vispruner25}, DART~\cite{dart25},  DivPrune~\cite{DivPrune25} and HoloV~\cite{holov25}. Each method offers a unique perspective on balancing computational cost and model performance. 

\smallskip
\noindent \textbf{Implementation Details.} 
\rev{We identify high-norm outliers with the scale-free top-$\rho$ rule of Sec.~\ref{sec:motivation} and keep $\rho=1\%$ for every backbone and every benchmark, so no per-model threshold calibration is involved.} For low-norm selection, we configure both the salience and diversity pool sizes to match the target token budget (e.g., 64 for \textit{Retain-64} and 32 for \textit{Retain-32}). More experimental settings and baseline details are provided in the Appendix~\ref{app:impl_details}.

% \smallskip
% \noindent \textbf{Implementation details.} xx leave to Appendix

\subsection{Main Results}

\noindent\textbf{\method{} achieves superior effectiveness under aggressive pruning}
As shown in Tab.~\ref{tab:llava1-5}, we evaluate our proposed \textit{\method{}}\ on LLaVA-1.5-7B under two extremely challenging pruning configurations, retaining only 64 ($\downarrow88.9\%$) and 32 ($\downarrow94.4\%$) visual tokens. Following prior work~\cite{ToMe23,FastV24,MustDrop24,PDdrop24,SparseVLM25,VisionZip25,vispruner25,holov25}, we report all results in normalized percentage form, where the vanilla 576-token model is treated as the 100\% upper bound.
In the 64-token setting, \textit{\method{}}\ attains an average performance of \textbf{96.5\%}, reducing nearly 90\% of visual tokens while incurring only a 3.5\% drop from the full model. It surpasses strong recent methods: exceeding VisionZip by +3.3\% and HoloV by +1.8\%. 
When retaining only 32 visual tokens, \textit{\method{}}\ continues to deliver strong performance with an average of \textbf{91.2\%}, outperforming all existing pruning methods by a clear margin. Compared with the best prior method VisPruner, \textit{\method{}}\ achieves a +4.0\% improvement, and exceeds VisionZip by +6.0\%. Even in this extremely low-token regime, our method consistently achieves the top performance across all benchmarks, showing that \textit{\method{}}\ maintains high fidelity even when over 94\% of the visual tokens are removed.

% These results collectively demonstrate that \textit{\method{}} not only preserves strong image understanding capability under moderate pruning but also remains remarkably stable under extreme compression, highlighting its effectiveness and generality as a visual token reduction framework for LLaVA-1.5-7B.

\smallskip
\noindent\textbf{\method{} generalizes seamlessly to dynamic-resolution architectures.}
To further demonstrate the generality of \textit{\method{}}, we evaluate its performance on Qwen2.5-VL-7B~\cite{qwen25}, a state-of-the-art MLLM whose visual pipeline differs fundamentally from the LLaVA family. Unlike LLaVA models that rely on a fixed-grid ViT encoder (\eg, $336 \times 336$), Qwen2.5-VL-7B employs a \textit{Naive Dynamic Resolution} mechanism. This architecture allows the model to process images of arbitrary aspect ratios by mapping variable pixel counts to a dynamic number of visual tokens. While this flexibility enhances fine-grained perception for high-resolution inputs, it often results in massive token sequences, leading to significant computational overhead and latent redundancy. Pruning tokens in such a variable-length regime is more challenging, as the model lacks a stable spatial prior. \rev{Because this vision tower has no [\texttt{CLS}] token, we instantiate the \texttt{CLS}-free salience score introduced in Sec.~\ref{sec:method}, which makes \textit{\method{}} directly applicable without architectural changes.} Our results show that even within this dynamic resolution framework, \textit{\method{}} consistently achieves the best average performance, proving its robustness across diverse MLLM architectures.
Following the experimental settings in HoloV~\cite{holov25}, we assess performance under aggressive pruning ratios of 66.7\%, 77.8\%, and 88.9\%. As detailed in Tab.~\ref{tab:qwen_main}, \textit{\method{}} demonstrates remarkable robustness, retaining \textbf{98.6\%}, \textbf{96.3\%}, and \textbf{91.8\%} of the full-model performance, respectively. In contrast, existing SOTA methods (\eg, FastV~\cite{FastV24}, HoloV~\cite{holov25}\rev{, VisionZip~\cite{VisionZip25}, DivPrune~\cite{DivPrune25}, MMTok~\cite{mmtok26}}) degrade sharply in this dynamic regime, confirming \textit{\method{}}'s superiority in identifying redundancy even within complex, variable-length visual sequences.

\smallskip
\noindent\rev{\textbf{\method{} preserves performance on harder reasoning tasks.}}
\rev{The benchmarks above are largely perception-oriented, so a natural concern is whether pruning preserves perception while destroying the evidence needed for multi-step reasoning. We therefore evaluate on the four harder benchmarks of Sec.~\ref{sec:exp}. As shown in Tab.~\ref{tab:harder_bench}, under the most aggressive 32-token budget \textit{\method{}} retains \textbf{93.7\%} of the full-model average, versus 88.7\% for VisionZip and 85.9\% for FastV. The gap is largest on MM-Vet (29.36 vs.\ 25.23 and 22.66) and AI2D (53.63 vs.\ 51.81 and 50.49), \ie, precisely the tasks requiring several spatially distinct pieces of evidence, while on MMMU all three stay within 0.6 points since that benchmark is dominated by textual knowledge. Sanitizing high-norm tokens therefore does not trade reasoning fidelity for perception fidelity. Results on LLaVA-1.5-13B and the high-resolution LLaVA-NeXT-7B, which verify transfer across model scale and visual-token regimes, are in Appendix~\ref{app:generalization}.}

\smallskip
\noindent\textbf{\method{} exhibits robust cross-modal transferability to videos.} 
To demonstrate the generalizability of \textit{\method{}} on video tasks, we evaluate \textit{\method{}} on Qwen2.5-VL-7B. As shown in Tab.~\ref{tab:video_qwen25}, even under an aggressive \textbf{80\% pruning ratio}, \textit{\method{}} achieves an impressive average performance retention of \textbf{98.0\%}. It consistently outperforms recent SOTA video pruning methods like DART~\cite{dart25} and DivPrune~\cite{DivPrune25}, validating its robustness in capturing spatiotemporal cues without any retraining.

\begin{table}[t]
    \centering
    \setlength{\tabcolsep}{2pt}
    \tiny

    \begin{tabular}{l | *{4}{c}}
    \toprule
        \textbf{Methods} & \textbf{Time (mm:ss)} & \textbf{Prefill (ms)} & \textbf{Latency (ms)} & \textbf{Acc.} \\
    \midrule
        
        \textcolor{gray}{Upper Bound (576 Tokens)} 
            & \textcolor{gray}{19:28}
            & \textcolor{gray}{62.75}
            & \textcolor{gray}{121.0}
            & \textcolor{gray}{100\%} \\
        \midrule
        
        \rowcolor{mygray}
        LLaVA-1.5-7B 
            & \multicolumn{4}{c}{\textit{Pruning Ratio = $\fg{\downarrow 90.0\%}$}} \\
        
        FastV \texttt{\scriptsize{(ECCV24)}} 
            & 12:02 & 32.2 & 75.0 & 65.5\% \\
        
        VisionZip \texttt{\scriptsize{(CVPR25)}} 
            & 12:24 & 36.9 & 77.4 & 91.8\% \\
        
        SparseVLM \texttt{\scriptsize{(ICML25)}} 
            & 12:20 & 38.6 & 77.0 & 90.5\% \\
        
        \rowcolor{mygreen2}
        \textbf{\method{} \scriptsize{(Ours)}}
            & 12:59 & 37.1 & 86.0 & 97.1\% \\
    \bottomrule
	\end{tabular}
    \caption{Real inference comparison on POPE~\cite{pope23}. Experiments adopt 90\% pruning ratio.}
    \label{tab:efficiency}
\end{table}

% \begin{table}[t]
%     \centering
%     \setlength{\tabcolsep}{4.5pt}
%     \footnotesize
%     \caption{Real inference comparison on POPE. Experiments adopt 90\% pruning ratio.}
%     \label{tab:efficiency}
%     \vspace{0.25em}
%     \resizebox{0.70\linewidth}{!}{
%     \begin{tabular}{l | *{5}{>{\centering\arraybackslash}p{1.05cm}}}
%         \textbf{Methods} & \textbf{Time} & \textbf{Prefill (ms)} & \textbf{Latency (ms)} & \textbf{Mem.} & \textbf{Rel. Acc.} \\
%         \midrule
        
%         \textcolor{gray}{Upper Bound (576 Tokens)} 
%             & \textcolor{gray}{19:28}
%             & \textcolor{gray}{62.75}
%             & \textcolor{gray}{121.0}
%             & \textcolor{gray}{19.0G}
%             & \textcolor{gray}{100\%} \\
%         \midrule
        
%         \rowcolor{mygray}
%         LLaVA-1.5-7B 
%             & \multicolumn{5}{c}{\textit{Token Pruning Rate = $\fg{\downarrow 90.0\%}$}} \\
        
%         FastV \texttt{\scriptsize{(ECCV24)}} 
%             & 12:02 & 32.2 & 75.0 & 15.7G & 65.5\% \\
        
%         VisionZip \texttt{\scriptsize{(CVPR25)}} 
%             & 12:24 & 36.9 & 77.4 & 14.6G & 91.8\% \\
        
%         SparseVLM \texttt{\scriptsize{(ICML25)}} 
%             & 12:20 & 38.6 & 77.0 & 14.5G & 90.5\% \\
        
%         \rowcolor{mygreen2}
%         \method{} \scriptsize{(Ours)}
%             & 12:59 & 37.1 & 86.0 & 14.6G & 97.1\% \\
% 	\end{tabular}}
%     \vspace{-1em}
% \end{table}

We evaluate efficiency on LLaVA-1.5 (Tab.~\ref{tab:efficiency}). With 90\% token pruning, \textit{\method{}} reduces total inference time by \textbf{33.3\%} compared to the full model. While our cascading design introduces a marginal latency overhead compared to competitive methods like VisionZip, it yields a \textbf{commanding accuracy advantage} (97.1\% vs. 91.8\%). Unlike FastV which suffers catastrophic degradation (65.5\%), \textit{\method{}} achieves the optimal practical trade-off, delivering substantial acceleration without compromising reasoning reliability.

\subsection{Ablation Studies}
\label{subsec:ablation}

\begin{table}[t]
\centering
\scriptsize
\setlength{\tabcolsep}{3pt}

\begin{tabular}{lcc}
\toprule
\textbf{Method} & \textbf{MME $\uparrow$} & \textbf{MMB $\uparrow$} \\
\midrule
w/o Visual Sanitizer & 1589.0 & 51.4 \\
w/o Text-Guided Pruner & 1690.0 & 59.4 \\
w/o Removing Duplicates & 1733.3 & 60.1 \\
w/o Removing High-Norm & 1705.6 & 59.9 \\
w/o High-Norm Aggregation & 1737.0 & 60.2 \\
\midrule
\rowcolor{gray!10} \textbf{\method{}} & \textbf{1754.1} & \textbf{61.6} \\
\bottomrule
\end{tabular}
\caption{\textbf{Ablation studies on components.} We report performances on MME(~\cite{mme23} and MMBench(~\cite{mmbench24}).}
\label{tab:ablation}
\vspace{-10pt}
\end{table}
To validate the effectiveness of \textit{\method{}}, we dissect the contribution of each component in Tab.~\ref{tab:ablation}. 
The most significant finding is the criticality of the pre-filtering stage: removing the \textit{visual sanitizer} entirely causes a catastrophic performance drop (\eg, \textbf{-10.2\%} on MMB). This strongly validates our core premise that without mitigating inherent visual redundancy upstream, the LLM is overwhelmed by attention sinks and dispersion, rendering subsequent reasoning ineffective. Similarly, omitting the \textit{text-guided pruner} leads to a notable decline, confirming that semantic alignment is indispensable for fine-grained understanding.
Inside the \textit{visual sanitizer}, the \textit{high-norm removal} proves to be the dominant factor. Retaining these high-norm outlier tokens causes a sharp degradation (MME: $1754.1 \to 1705.6$), empirically verifying that high-norm outlier tokens are indeed redundant. In contrast, removing similarity-based de-duplication yields a smaller drop, indicating that while spatial diversity matters, the primary gain stems from eliminating the high-norm outlier tokens that disrupt the attention landscape. Notably, \textit{high-norm aggregation} performs slightly better than direct removal, suggesting that preserving a compact summary of these tokens is marginally more effective than discarding them entirely. \rev{Additional sensitivity analysis on the high-norm criterion is provided in Appendix~\ref{app:sensitivity_analysis}, and Appendix~\ref{app:hyper_ablation} further ablates the remaining design choices---the pruning layers, the progressive retention schedules, and the salience-to-diversity split---under matched token-computation budgets. Both analyses show flat response surfaces, indicating that our reported configuration is a simple benchmark-independent default rather than a tuned optimum.}

% \subsection{Transferability to Existing Pruning Frameworks}
\subsection{Transferability to Existing Methods}
\label{subsec:Transferability}

\begin{table}[t]
    \centering
    \scriptsize
    \setlength{\tabcolsep}{1.8pt}

    \begin{tabular}{@{}p{3.2cm}c cc @{\hspace{4pt}} cc@{}}
    \toprule
    & & \multicolumn{2}{c}{\textbf{MMB}} & \multicolumn{2}{c}{\textbf{POPE}} \\
    \cmidrule{3-4} \cmidrule{5-6}
    \textbf{Method} & \textbf{Tok} & \textbf{Acc} & \textbf{Rel.} & \textbf{Acc} & \textbf{Rel.} \\
    \midrule
    
    % NOTE: the Vanilla MMB score is 64.7, matching the Upper Bound row of
    % Tab.~\ref{tab:llava1-5}; the Rel. column is recomputed against it.
    \mbox{Vanilla (Full)} & 576 & \rev{64.7} & 100\% & 85.9 & 100\% \\

    \midrule

    \mbox{VisionZip~\cite{VisionZip25}} & 64 & 60.1 & \rev{92.9\%} & 77.0 & 89.6\% \\
    \rowcolor{gray!10} \mbox{+ High-Norm Filter} & 64 & \textbf{61.4} & \textbf{\rev{94.9\%}} & \textbf{77.7} & \textbf{90.5\%} \\

    \midrule

    \mbox{VisionZip~\cite{VisionZip25}} & 32 & 57.7 & \rev{89.2\%} & 68.7 & 80.0\% \\
    \rowcolor{gray!10} \mbox{+ High-Norm Filter} & 32 & \textbf{59.4} & \textbf{\rev{91.8\%}} & \textbf{70.4} & \textbf{82.0\%} \\
    
    \bottomrule
    \end{tabular}
    \caption{
    \textbf{Transferability to existing methods.} Integrating our high-norm filtration module into VisionZip consistently yields performance gains.
    }
    \label{tab:generality_verification}
    \vspace{-10pt}
\end{table}
To further demonstrate that our identification of high-norm outlier tokens is a fundamental insight rather than a method-specific heuristic, we evaluate the transferability of our proposed visual filtration module. Specifically, we investigate whether our ``high-norm removal'' strategy can serve as an orthogonal, plug-and-play enhancement for existing vision-centric pruning frameworks. 

We integrate our norm-based separation step into the VisionZip~\cite{VisionZip25} pipeline, applying it before its standard dominant token selection. As shown in Tab.~\ref{tab:generality_verification}, at identical token reduction ratios (32 and 64 tokens), the inclusion of our high-norm outliers filter yields a consistent and significant performance boost ranging from \textbf{0.9\% to 2.6\%} across both MMBench and POPE. 

This improvement is particularly noteworthy because VisionZip, as a state-of-the-art attention-based method, inherently prioritizes tokens with high \textit{\texttt{CLS} attention}---the exact group we identified as high-norm outlier tokens. By preemptively removing these outliers, we allow the baseline's selection mechanism to focus on truly informative low-norm tokens. These results validate that our proposed filtration module is highly versatile and can be seamlessly generalized to enhance other vision-centric strategies.

% \subsection{Sensitivity Analysis}
% \label{app:sensitivity_analysis}
% \input{table/tau_sensitve_analysis}
% \textcolor{green}{
% We further study the sensitivity of the high-norm threshold $\tau$, which is used to identify high-norm outlier tokens in the visual sanitizer.}
% \textcolor{green}{
% As shown in Tab.~\ref{tab:tau_sensitivity}, the performance remains stable across a relatively broad range of threshold values once $\tau$ is sufficiently large. In particular, choosing $\tau$ in the range of $[45,90]$ leads to only minor variations on both POPE and MME, indicating that our method is not overly sensitive to the exact choice of the threshold.
% In contrast, an excessively small threshold ($\tau=30$) causes a clear performance drop, suggesting that overly aggressive filtering may remove useful visual information together with true outliers.
% These results support our default choice of $\tau=60$ as a robust operating point.} 

% \input{sec/6_analysis}
\section{Conclusion}

In this work, we presented \textit{\method{}}, a training-free cascading visual token pruning framework for efficient MLLM inference. By filtering high-norm outlier tokens with a \textit{visual sanitizer}, our method suppresses massive activations and attention dispersion in the LLM decoder, thereby enabling more precise \textit{text-guided pruning}.

Extensive experiments on diverse image and video benchmarks demonstrate that \textit{\method{}} achieves aggressive token reduction while preserving strong perception and reasoning performance. We further show that our framework generalizes well across different MLLM architectures, from fixed-grid to dynamic-resolution models. Finally, the transferability of the \textit{visual sanitizer} as a plug-and-play module highlights its potential to improve existing pruning methods more broadly.

% \vspace{0.5em}
% \noindent\textbf{Limitations.} While effective for static images and finite video clips, our current evaluation focuses on offline inference with fixed-length inputs. Real-world edge applications, such as continuous robotic perception, often require handling infinite video streams with evolving temporal contexts. Adapting our coarse-to-fine pruning strategy to support online streaming---where past visual history is dynamically updated without future knowledge---remains an open challenge.

% \section*{Limitations}
\smallskip
\noindent\textbf{Limitations.}
While effective for static images and finite video clips, our current evaluation focuses solely on offline inference with pre-recorded, fixed-length inputs available in advance. Real-world edge applications, such as continuous robotic perception, often require handling potentially infinite video streams with continuously evolving temporal contexts. Adapting our coarse-to-fine pruning strategy to support online streaming---where past visual history is dynamically updated without access to any future knowledge, and where pruning decisions cannot be revisited---remains an open challenge for future work.

%\smallskip
%\noindent\textbf{Acknowledgements.}
%This work was supported by 

% -----------------------------------------------------------------------------
% Optional sections required/encouraged by ACL-family venues
% -----------------------------------------------------------------------------
% Uncomment and fill these if required by the EMNLP submission instructions.
% \section*{Limitations}
% \input{sec/limitations}
%
% \section*{Acknowledgments}
% \input{sec/acknowledgments}

% -----------------------------------------------------------------------------
% Bibliography
% -----------------------------------------------------------------------------
% ACL style automatically uses acl_natbib.bst through acl.sty.
% Your ECCV main.tex used main.bib, so we keep the same .bib database here.
\bibliography{main}

% -----------------------------------------------------------------------------
% Appendix
% -----------------------------------------------------------------------------
% Uncomment this if you have appendix content prepared.
\appendix

% \appendix
% --- 新增的 Appendix 简介大纲 ---
\noindent In this appendix, we provide further details beyond the main paper, including further validation of the norm distribution on non-CLIP visual encoders (Sec.~\ref{app:nonclip_norm}), sensitivity analysis of the high-norm threshold (Sec.~\ref{app:sensitivity_analysis}), additional implementation details (Sec.~\ref{app:exp_details} and Sec.~\ref{app:impl_details}), additional experimental results on diverse image and video benchmarks (Sec.~\ref{app:additional_exps}), \rev{generalization across model scales and visual-token regimes (Sec.~\ref{app:generalization}), systematic ablations of the remaining hyperparameters (Sec.~\ref{app:hyper_ablation}),} and additional visualizations of attention outliers and spatial redundancy (Sec.~\ref{app:more_visuals}).

\section{Experiments on Non-CLIP Visual Encoders}
\label{app:nonclip_norm}

\begin{figure}[t]
    \centering
    \includegraphics[width=0.5\linewidth]{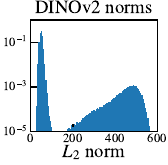}
    \caption{\textbf{Norm distributions of non-CLIP visual encoders.}
    We visualize the patch feature norm distributions of DINO and DINOv2.
    Both encoders show a small set of high-norm outliers separated from the majority of regular tokens.
    For DINOv2, although most patch tokens lie below 100, 2.37\% of tokens have norms larger than 150, indicating that the high-norm outlier phenomenon generalizes beyond CLIP-based encoders.
    \rev{The differing norm scales motivate our scale-free top-$\rho$ criterion in place of an absolute threshold.}}
    \label{fig:dinov2_norm}
\end{figure}

To verify that the high-norm outlier phenomenon is not unique to CLIP-based visual encoders, we further analyze the feature norm distributions of non-CLIP encoder, namely DINOv2.
As shown in Fig.~\ref{fig:dinov2_norm}, both encoders exhibit a clear separation between the majority of regular patch tokens and a small group of high-norm outliers.
This suggests that the existence of abnormal high-norm tokens is a broader property of modern vision encoders, rather than an artifact specific to CLIP.

For DINOv2 in particular, although most patch tokens have $\ell_2$ norms roughly within the range $[0,100]$, a small proportion of tokens have substantially larger norms.
In our measurement, the fraction of tokens with norm larger than 150 is 2.37\%.
This confirms that high-norm outliers remain sparse, but sufficiently prominent to distort the feature distribution and potentially interfere with downstream token compression.

At the same time, the exact norm scale differs across encoders.
Compared with CLIP-based features used in our main experiments, DINOv2 exhibits a substantially shifted norm range, with its outlier mode appearing at much larger values.
\rev{This is precisely why \textit{\method{}} does not rely on an absolute cutoff. The top-$\rho$ rule of Sec.~\ref{sec:motivation} ranks tokens \emph{within} each encoder, so it adapts automatically to a shifted norm range and can be transferred across model families without any per-encoder calibration. An absolute threshold, in contrast, would have to be re-derived from the empirical norm distribution of every new target encoder.}

\section{Sensitivity Analysis}
\label{app:sensitivity_analysis}
\begin{table}[t]
\centering
\scriptsize
\setlength{\tabcolsep}{4.5pt}
\renewcommand{\arraystretch}{1.1}

\begin{tabular}{lccccc}
\toprule
\textbf{Benchmark} & \multicolumn{5}{c}{\textbf{Threshold }$\boldsymbol{\tau}$} \\
\cmidrule(lr){2-6}
& \textbf{30} & \textbf{45} & \cellcolor{gray!10}\textbf{60} & \textbf{75} & \textbf{90} \\
\midrule
POPE $\uparrow$ & 10.7  & 77.8 & \cellcolor{gray!10}78.5 & 78.3 & 77.5 \\
MME $\uparrow$  & 752  & 1324 & \cellcolor{gray!10}1344 & 1334 & 1326 \\
\bottomrule
\end{tabular}

\caption{
\textbf{Sensitivity analysis of the high-norm threshold $\tau$.}
Performance remains stable across a broad range of threshold values, while an overly small threshold degrades performance.
}
\label{tab:tau_sensitivity}
\end{table}

We further study the sensitivity of the high-norm criterion used to identify high-norm outlier tokens in the visual sanitizer, \rev{by sweeping an absolute norm threshold $\tau$ on LLaVA-1.5}.

As shown in Tab.~\ref{tab:tau_sensitivity}, the performance remains stable across a relatively broad range of threshold values once $\tau$ is sufficiently large. In particular, choosing $\tau$ in the range of $[45,90]$ leads to only minor variations on both POPE and MME, indicating that our method is not overly sensitive to the exact choice of the threshold.
In contrast, an excessively small threshold ($\tau=30$) causes a clear performance drop, suggesting that overly aggressive filtering may remove useful visual information together with true outliers.
\rev{The practical message is that high-norm filtering only requires separating the small high-norm tail from the bulk of the distribution, and does not require a precisely tuned operating point. Our reported experiments therefore use the equivalent but scale-free top-$\rho$ rule with $\rho=1\%$, which realizes this separation without exposing an encoder-dependent constant.}

\section{Experimental Details}
\label{app:exp_details}

\subsection{Benchmarks and Metrics}
\label{app:benchmarks_metrics}

We evaluate our method on a diverse set of widely used benchmarks covering both image-language and video-language understanding.
For \textbf{image-based} tasks, we report results on GQA~\cite{GQA19}, MMBench (MMB) and MMB-CN~\cite{mmbench24}, MME~\cite{mme23}, POPE~\cite{pope23}, ScienceQA (SQA)~\cite{scienceqa}, VQA-v2~\cite{balanced_vqa_v2}, and TextVQA~\cite{textvqa19}\rev{, together with four harder reasoning-oriented benchmarks: MMStar~\cite{mmstar24}, MMMU~\cite{mmmu24}, AI2D~\cite{ai2d16}, and MM-Vet~\cite{mmvet24}}.
For \textbf{video-based} tasks, we additionally evaluate on NextQA~\cite{nextqa21}, MVBench~\cite{mvbench}, SEED-Bench~\cite{seed_bench23}, and VideoMME~\cite{videomme}. The detailed information of these tasks is listed below.

% \smallskip
% \noindent \textbf{Image benchmarks.}

\smallskip
\noindent \textbf{GQA}~\cite{GQA19}.
GQA is a benchmark for real-world visual reasoning and compositional question answering.
Its questions are grounded in scene graph structures and paired with functional programs, which makes it well suited for evaluating grounded reasoning, relational understanding, and robustness to superficial language priors.

\smallskip
\noindent \textbf{MMBench / MMB-CN}~\cite{mmbench24}.
MMBench is a systematically designed multiple-choice benchmark for holistic evaluation of vision-language models.
It organizes evaluation into hierarchical ability dimensions spanning perception and reasoning, while MMB-CN provides a Chinese counterpart that enables bilingual and multilingual assessment under a unified protocol.

\smallskip
\noindent \textbf{MME}~\cite{mme23}.
MME is a comprehensive benchmark for multimodal large language models that measures both perception and cognition abilities.
It consists of 14 subtasks with manually designed instruction-answer pairs and concise prompts, making evaluation more controlled and less sensitive to prompt engineering.

\smallskip
\noindent \textbf{POPE}~\cite{pope23}.
POPE focuses on object hallucination in large vision-language models.
It reformulates hallucination evaluation as a set of binary probing questions about object presence and reports metrics such as Accuracy, Precision, Recall, and F1 under multiple sampling strategies, providing a direct measure of hallucination behavior.

\smallskip
\noindent \textbf{ScienceQA}~\cite{scienceqa}.
ScienceQA is a multimodal science question answering benchmark with about 21K multiple-choice questions spanning diverse science topics.
In addition to answers, it provides lectures and explanations, so it evaluates not only perception but also multi-step scientific reasoning across text and image inputs.

\smallskip
\noindent \textbf{VQA-v2}~\cite{balanced_vqa_v2}.
VQA-v2 is a large-scale open-ended visual question answering benchmark built on diverse real-world images.
Each question is paired with ten human answers and evaluated with the standard VQA metric, making it a standard testbed for general visual perception, commonsense grounding, and answer robustness.

\smallskip
\noindent \textbf{TextVQA}~\cite{textvqa19}.
TextVQA evaluates a model's ability to answer questions that require reading and reasoning over scene text embedded in images.
Compared with standard VQA, it places greater emphasis on OCR-sensitive perception and fine-grained integration of textual and visual evidence.

\smallskip
\noindent \rev{\textbf{MMStar}~\cite{mmstar24}.}
\rev{MMStar is a vision-indispensable benchmark curated to reduce language-prior leakage, \ie, questions that can be answered from text alone.
It therefore provides a stricter measure of whether a model genuinely grounds its reasoning in the visual input.}

\smallskip
\noindent \rev{\textbf{MMMU}~\cite{mmmu24}.}
\rev{MMMU is a massive multi-discipline benchmark of college-level problems spanning science, engineering, and the humanities.
It requires knowledge-intensive, expert-level reasoning over heterogeneous visual formats such as charts, chemical structures, and diagrams. We report results on its validation split.}

\smallskip
\noindent \rev{\textbf{AI2D}~\cite{ai2d16}.}
\rev{AI2D is a diagram understanding benchmark built from grade-school science diagrams.
Answering its questions requires parsing structured visual layouts, including arrows, labels, and part-whole relations, rather than recognizing natural objects.}

\smallskip
\noindent \rev{\textbf{MM-Vet}~\cite{mmvet24}.}
\rev{MM-Vet evaluates integrated multimodal capabilities through open-ended questions graded by an LLM judge.
Because each question typically requires composing several skills, such as recognition, OCR, spatial reasoning, and knowledge retrieval, it is sensitive to the loss of any single piece of visual evidence.}

% \smallskip
% \noindent \textbf{Video benchmarks.}

\smallskip
\noindent \textbf{NextQA}~\cite{nextqa21}.
NextQA is a video question answering benchmark designed to move beyond descriptive QA toward explaining temporal actions.
It emphasizes causal and temporal reasoning in realistic videos, and many questions require tracking events, object interactions, and temporal dependencies across multiple frames.

\smallskip
\noindent \textbf{MVBench}~\cite{mvbench}.
MVBench is a comprehensive benchmark for video understanding that targets temporal reasoning abilities of multimodal models.
It covers 20 challenging tasks that cannot be solved reliably from a single frame, spanning a broad range of temporal skills from low-level perception to higher-level cognition.

\smallskip
\noindent \textbf{SEED-Bench}~\cite{seed_bench23}.
SEED-Bench is a large-scale multiple-choice benchmark for multimodal large language models with accurate human annotations.
It spans a broad set of evaluation dimensions and includes both image- and video-centric scenarios, offering an efficient and relatively objective way to compare general multimodal capability.

\smallskip
\noindent \textbf{VideoMME}~\cite{videomme}.
VideoMME is a comprehensive benchmark for evaluating multimodal large language models on video analysis.
It is designed to assess full-spectrum video understanding, with an emphasis on temporal comprehension, compositional reasoning, and robust performance across diverse real-world video scenarios.

\smallskip
\noindent \textbf{Metrics.}
We follow the official evaluation protocols of each benchmark.
For classification-style benchmarks such as MMBench, we report \textbf{accuracy}.
For VQA-style benchmarks such as VQA-v2, TextVQA, and NextQA, we report the \textbf{official VQA score/accuracy} computed by the released evaluation scripts.
For MME, we report the \textbf{standard MME score} under its official setup.
For POPE, we report \textbf{Accuracy}, and additionally Precision/Recall/F1 when required by the benchmark.

\subsection{Comparison methods.} We compare \textit{\method{}} with existing state-of-the-art (SOTA) visual token reduction methods based on different MLLM architectures (\ie, LLaVA-1.5~\cite{llava1_5} and Qwen2.5-VL~\cite{qwen25}). These baselines employ diverse strategies---such as token merging, attention-based pruning, adaptive allocation, and hierarchical retention---to improve efficiency by reducing redundant tokens. Specifically, we compare against ToMe~\cite{ToMe23}, MustDrop~\cite{MustDrop24}, FastV~\cite{FastV24}, PDrop~\cite{PDdrop24}, PruMerge~\cite{prumerge24}, SparseVLM~\cite{SparseVLM25}, VisionZip~\cite{VisionZip25}, Vispruner~\cite{vispruner25}, DART~\cite{dart25},  DivPrune~\cite{DivPrune25}, HoloV~\cite{holov25}\rev{, MMTok~\cite{mmtok26}} and ApET~\cite{ma2026apet}. Each method offers a unique perspective on balancing computational cost and model performance. 
\rev{For the dynamic-resolution Qwen2.5-VL experiments, VisionZip, DivPrune, and MMTok are re-implemented on top of the same \texttt{lmms-eval} harness and the same \texttt{max\_pixels} setting used for \textit{\method{}}, so that all methods observe an identical visual token sequence before pruning.}
% \subsection{Training-Free / Inference Settings}
% \label{app:inference_settings}
% % TODO: Token budgets, pruning ratios, decoding params (temperature, top-p, etc.).

\section{Implementation Details}
\label{app:impl_details}
\subsection{LLaVA Implementation}
\label{app:impl_llava}

\noindent \textbf{Backbone and evaluation.}
For all experiments on the LLaVA family, we use \texttt{LLaVA-1.5-7B} as the base model and follow the official evaluation pipeline released by LLaVA.
The original visual input contains a fixed number of 576 image tokens.
We adopt the corresponding official toolkit or evaluation server for each benchmark whenever available.

\smallskip
\noindent \textbf{Visual sanitizer.}
\rev{We identify high-norm outliers with the scale-free top-$\rho$ rule, using $\rho = 1\%$ of the visual tokens.}
For low-norm token selection, we set both the salience pool size and the diversity pool size to match the target retained token budget.
Specifically, the two pool sizes are both set to 128 for \textit{Retain-128}, 64 for \textit{Retain-64}, and 32 for \textit{Retain-32}.

\smallskip
\noindent \textbf{Text-guided pruner.}
For LLaVA-1.5-7B, we perform text-guided pruning at three layers of the language model, namely layers 2, 6, and 15, corresponding to an early, middle, and late stage of multimodal decoding.
At each pruning stage, the number of visual tokens is further reduced in a progressive manner.
For the three target budgets, the retained visual tokens after the three pruning stages are set to $(230, 130, 92)$, $(110, 74, 42)$, and $(54, 36, 22)$, respectively.
This progressive schedule allows the model to gradually remove redundant tokens while preserving important visual evidence for later reasoning.

\smallskip
\noindent \textbf{Pruning configurations.}
Tab.~\ref{tab:llava_pruning_config} summarizes the pruning hyperparameters used in our LLaVA experiments.

\begin{table}[t]
    \centering
    \caption{\textbf{Pruning hyperparameters for LLaVA-1.5-7B under different target token budgets.}
    The original visual input contains 576 tokens.
    For low-norm selection, both the salience pool and the diversity pool are set equal to the target retained token budget.
    \rev{The high-norm fraction $\rho$ is shared by all budgets and all backbones.}}
    \label{tab:llava_pruning_config}
    \resizebox{0.7\linewidth}{!}{
    \begin{tabular}{c|c|c|c|c}
        \toprule
        \textbf{Target Budget} & \boldmath$\rho$ & \textbf{Pruning Layers} & \textbf{low-norm Pools} & \textbf{Progressive Token Schedule} \\
        \midrule
        Retain-128 & 1\% & (2, 6, 15) & $(128, 128)$ & $(230, 130, 92)$ \\
        Retain-64  & 1\% & (2, 6, 15) & $(64, 64)$   & $(110, 74, 42)$ \\
        Retain-32  & 1\% & (2, 6, 15) & $(32, 32)$   & $(54, 36, 22)$ \\
        \bottomrule
    \end{tabular}}
\end{table}

\subsection{Qwen-VL Implementation}
\label{app:impl_qwen}

\noindent \textbf{Backbone and evaluation.}
For all experiments on the Qwen family, we use \texttt{Qwen2.5-VL} as the base model and conduct evaluation with the \texttt{lmms-eval} framework.
We use the default frame sampling strategy of the original model.
The visual resolution is controlled by setting \texttt{max\_pixels=376320}.

\smallskip
\noindent \textbf{Visual sanitizer.}
For Qwen2.5-VL, we adopt the same visual sanitizer as in the LLaVA setting.
For low-norm token selection, we divide the candidate tokens into a \emph{salience pool} and a \emph{diversity pool}, following the same terminology used in the main paper.
The diversity pool is always set to $5\%$ of the current visual tokens (with at least one token kept), while the salience pool is adjusted according to the target pruning ratio.
Specifically, the salience pool is set to $45\%$, $45\%$, and $20\%$ of the current visual tokens for the 66.7\%, 77.8\%, and 88.9\% pruning settings, respectively.
\rev{Since this vision tower provides no [\texttt{CLS}] token, the salience term is instantiated by the \texttt{CLS}-free score described in Sec.~\ref{app:cls_free}.}

\smallskip
\noindent \textbf{Text-guided pruner.}
We apply progressive text-guided pruning in three stages.
For different target pruning ratios, the three stages use different retention schedules to gradually reduce the visual tokens.
When the target pruning ratio is $66.7\%$, the three stages retain $40\%$, $35\%$, and $25\%$ of the current tokens, respectively.
When the target pruning ratio is $77.8\%$, the corresponding stage-wise retention ratios are $30\%$, $20\%$, and $12\%$.
For the most aggressive setting with a target pruning ratio of $88.9\%$, the three stages retain $18\%$, $11\%$, and $6\%$ of the current tokens, respectively.
This progressive design enables the model to remove redundant visual tokens step by step while preserving useful evidence for subsequent multimodal reasoning.

\smallskip
\noindent \textbf{Pruning configurations.}
Tab.~\ref{tab:qwen_pruning_config} summarizes the pruning hyperparameters used in our Qwen2.5-VL experiments.

\begin{table}[t]
    \centering
    \caption{\textbf{Pruning hyperparameters for Qwen2.5-VL under different pruning ratios.}
    The diversity pool is always set to $5\%$ of the current tokens, while the salience pool is adjusted according to the target pruning ratio.}
    \label{tab:qwen_pruning_config}
    \resizebox{0.7\linewidth}{!}{
    \begin{tabular}{c|c|c|c}
        \toprule
        \textbf{Target Pruning Ratio} & \textbf{Salience Pool} & \textbf{Diversity Pool} & \textbf{Stage-wise Retention Ratios} \\
        \midrule
        66.7\% & 45\% & 5\% & (40\%, 35\%, 25\%) \\
        77.8\% & 45\% & 5\% & (30\%, 20\%, 12\%) \\
        88.9\% & 20\% & 5\% & (18\%, 11\%, 6\%) \\
        \bottomrule
    \end{tabular}}
\end{table}

The pool sizes are defined as proportions rather than fixed counts, since Qwen2.5-VL uses dynamic visual tokenization.

\subsection{Salience Scoring for Encoders without a \texttt{CLS} Token}
\label{app:cls_free}

\rev{The salience score $\mathcal{A}_{cls}$ used in Sec.~\ref{sec:method} presumes a [\texttt{CLS}]-based encoder such as CLIP-ViT, where the attention from the [\texttt{CLS}] token to each patch provides a natural global importance signal. Several modern vision towers, including that of Qwen2.5-VL~\cite{qwen25}, do not maintain such a token. We emphasize that \textit{\method{}} does not fundamentally depend on it: high-norm identification, high-norm aggregation, and diversity-based selection are all defined purely over patch features and are therefore already \texttt{CLS}-free. Only the salience term requires an equivalent substitute.}

\rev{For non-\texttt{CLS} encoders, we define the salience of visual token $j$ as its average \emph{received} visual self-attention,}
\begin{equation}
    s_j = \frac{1}{HN}\sum_{h=1}^{H}\sum_{i=1}^{N} A_{ij}^{(h)},
    \label{eq:cls_free_salience}
\end{equation}
\rev{where $A_{ij}^{(h)}$ is the attention weight from visual token $i$ to visual token $j$ in head $h$, $H$ is the number of attention heads, and $N$ is the number of visual tokens. Intuitively, $s_j$ measures how strongly the remaining visual context attends to token $j$, so it serves the same purpose as \texttt{CLS} attention: patches that many other patches rely on receive a high score. We compute Eq.~\eqref{eq:cls_free_salience} at the penultimate encoder layer, average over all heads, and substitute $s_j$ for $\mathcal{A}_{cls}$ in the top-$k_{res}$ selection; no other part of the pipeline changes.}

\rev{Our Qwen2.5-VL experiments validate this substitution. Under 88.9\% pruning, \textit{\method{}} with the \texttt{CLS}-free score retains 91.8\% of the full-model performance, exceeding HoloV by 5.6 points and VisionZip by 4.1 points (Tab.~\ref{tab:qwen_main}). This indicates that the received-attention formulation is a viable drop-in replacement rather than a degraded fallback.}

\subsection{Batched Diversity Selection}
\label{app:batched_selection}

\rev{Eq.~\eqref{eq:diversity} is written as a serial farthest-point rule for clarity: each round appends exactly one token, so obtaining $k_{div}$ diverse tokens requires $k_{div}$ sequential rounds. Because every round depends on the set selected so far, this loop cannot be parallelized on GPU and would dominate the prefill stage when the visual sequence is long, as in high-resolution or video inputs.}

\rev{Our implementation therefore uses a batched approximation. At round $t$, let $\mathcal{S}^{(t)}$ be the currently selected set and $\mathcal{R}^{(t)}$ the remaining candidates. We score all candidates simultaneously with a single matrix product, $d(x) = \max_{s \in \mathcal{S}^{(t)}} \text{CosSim}(x, s)$, and append the $b$ least similar ones at once:}
\begin{equation}
    \mathcal{B}^{(t)} = \mathrm{Bottom}_{b}\big(\{ d(x) \mid x \in \mathcal{R}^{(t)} \}\big),
    \label{eq:batched_diversity}
\end{equation}
\rev{where $\mathrm{Bottom}_{b}(\cdot)$ returns the $b$ candidates with the smallest scores. The two sets are then updated as $\mathcal{S}^{(t+1)} = \mathcal{S}^{(t)} \cup \mathcal{B}^{(t)}$ and $\mathcal{R}^{(t+1)} = \mathcal{R}^{(t)} \setminus \mathcal{B}^{(t)}$. We use $b=16$ throughout, which reduces the number of sequential rounds from $k_{div}$ to $\lceil k_{div}/16 \rceil$, \ie, by about $16\times$.}

\rev{Setting $b=1$ recovers Eq.~\eqref{eq:diversity} exactly. For $b>1$ the two rules differ, because the tokens within one batch are all scored against $\mathcal{S}^{(t)}$ and are therefore not penalized for being similar to each other. The batched variant is thus a close approximation rather than a mathematical equivalent of the serial rule. In practice the gap is small, since candidates are ranked by dissimilarity to an already diverse anchor set $\mathcal{S}^{(t)} \supseteq \mathbf{X}_{res}$, and $b$ is far smaller than $|\mathcal{R}^{(t)}|$. All numbers reported in this paper are produced with the batched variant.}

\subsection{Compute and Hardware}
\label{app:hardware}
% TODO: GPUs/CPUs, memory, runtime environment, and reproducibility notes.
% \noindent \textbf{Implementation Details.}
All experiments are conducted on NVIDIA A800-SXM4-80GB GPUs.
Our implementation is based on Python 3.10 with PyTorch 2.1.2 and CUDA 12.1.
For all benchmarks, we follow the official evaluation protocols and adopt the default settings of the corresponding baseline implementations.

\smallskip
\noindent\rev{\textbf{Attention implementation at pruning layers.}}
\rev{The text-guided pruner requires materialized text-to-vision attention scores, which standard FlashAttention kernels do not expose. We therefore fall back to an eager attention implementation at the three pruning layers only; every other layer keeps the efficient kernel and operates on an already shortened visual sequence. The efficiency numbers reported in Tab.~\ref{tab:efficiency} and Tab.~\ref{tab:efficiency_appendix} already include this overhead.}

\section{Additional Experiments}
\label{app:additional_exps}

In this section, we provide additional experimental results to complement the main paper.
Specifically, we report more comprehensive comparisons on both image and video benchmarks, together with additional real-inference results under different pruning ratios.
These results further verify that our method consistently achieves a stronger accuracy-efficiency trade-off across different architectures and evaluation settings.

For image understanding, we present extended benchmark results on LLaVA-1.5 in Tab.~\ref{tab:llava1-5}.
For video understanding, we provide additional comparisons on Qwen2.5-VL in Tab.~\ref{tab:video_appendix}.
We also include real inference measurements on POPE in Tab.~\ref{tab:efficiency_appendix}, reporting runtime, prefill cost, decoding latency, and relative accuracy under different pruning ratios.
Overall, these additional experiments further support the effectiveness and robustness of our method beyond the results shown in the main paper.

% \renewcommand{\multirowsetup}{\centering}
% \definecolor{mygray}{gray}{.92}
% \definecolor{mygreen1}{RGB}{253, 244, 244}
% \definecolor{mygreen2}{RGB}{238, 243, 243}
% \definecolor{ForestGreen}{RGB}{34,139,34}
% \newcommand{\fg}[1]{\mathbf{\mathcolor{ForestGreen}{#1}}}
% \definecolor{Forestred}{RGB}{220,50,50}
% \newcommand{\fr}[1]{\mathbf{\mathcolor{Forestred}{#1}}}

\begin{table*}
    \centering
    \setlength{\tabcolsep}{4.0pt}
    \scriptsize
    \resizebox{\linewidth}{!}{
    \begin{tabular}{l | *{8}{>{\centering\arraybackslash}p{1.05cm}} |>{\centering\arraybackslash}p{1.1cm}}
    \toprule
        \textbf{\;Methods} & \textbf{GQA} & \textbf{MMB} & \textbf{MMB}$_{\text{CN}}$ & \textbf{MME} & \textbf{POPE} & \textbf{SQA} & \textbf{VQA}$_{\text{V2}}$ & \textbf{VQA}$_{\text{Text}}$ & \makecell[c]{\textbf{Average}}\\
        \midrule
        
        % Upper bound
        \textcolor{gray}{Upper Bound, 576 Tokens} 
        & \textcolor{gray}{61.9} 
        & \textcolor{gray}{64.7} 
        & \textcolor{gray}{58.1} 
        & \textcolor{gray}{1862} 
        & \textcolor{gray}{85.9} 
        & \textcolor{gray}{69.5} 
        & \textcolor{gray}{78.4} 
        & \textcolor{gray}{58.2} 
        & \textcolor{gray}{100\%} \\
        \midrule

        % ================= Retain 128===================
        \midrule  \rowcolor{mygray}
        LLaVA-1.5 \textcolor{gray}{7B} & \multicolumn{8}{c}{\textit{Retain 128 Tokens} \ $\fg{(\text{Pruning Ratio}= 77.8\%)}$}\\
                
        ToMe \texttt{\scriptsize{(ICLR23)}} & 52.4 & 53.3 & - & 1343 & 62.8 & 59.6 & 63.0 & 49.1 & 80.4 \\
                
        FastV \texttt{\scriptsize{(ECCV24)}} & 49.6 & 56.1 & 56.4 & 1490 & 59.6 & 60.2 & 61.8 & 50.6 & 83.2 \\
                
        MustDrop \texttt{\scriptsize{(2024.11)}} & 56.9 & 61.1 & 55.2 & 1745 & 78.7 & 68.5 & 74.6 & 56.3 & 94.6 \\
                
        LLaVA-PruMerge \texttt{\scriptsize{(ICCV25)}} & 53.3 & 58.1 & 51.7 & 1554 & 67.2 & 67.1 & 68.8 & 54.3 & 88.0 \\
                
        PDrop \texttt{\scriptsize{(2024.10)}} & 56.0 & 61.1 & 56.6 & 1644 & 82.3 & 68.3 & 72.9 & 55.1 & 94.0 \\
                
        VisionZip \texttt{\scriptsize{(CVPR25)}} & 57.6 & 63.4 & 56.7 & 1768 & 84.7 & 68.8 & 75.6 & \textbf{56.8} & 96.9 \\
                
        SparseVLM \texttt{\scriptsize{(ICML25)}} & 56.0 & 60.0 & 51.1 & 1696 & 80.5 & 67.1 & 73.8 & 54.9 & 92.6 \\
                
        HoloV \texttt{\scriptsize{(NeurIPS25)}} & 57.7 & \textbf{63.9} & 56.5 & 1802 & 84.0 & \textbf{69.8} & 75.5 & \textbf{56.8} & 97.3 \\

        ApET \texttt{\scriptsize{(2026.02)}} 
        & 58.9 & 62.3 & 56.4 & 1801 & \textbf{86.1} & 68.7 & 75.1 & 53.9 & 97.0 \\
                
        \rowcolor{mygreen2}
        \textbf{\method{} (Ours)} 
        & \textbf{59.3} & 63.6 & \textbf{57.9} & \textbf{1806} & 85.3 & \textbf{69.8} & \textbf{76.8} & 55.8 & \textbf{98.0} \\
                
        \midrule

        % ================= Retain 64 ===================
        \rowcolor{mygray}
        LLaVA-1.5-7B & \multicolumn{8}{c}{\textit{Retain 64 Tokens} \ $\fg{(\text{Pruning Ratio}= 88.9\%)}$}\\

        ToMe \texttt{\scriptsize{(ICLR23)}} 
        & 48.6 & 43.7 & - & 1138 & 52.5 & 50.0 & 57.1 & 45.3 & 70.1 \\

        FastV \texttt{\scriptsize{(ECCV24)}} 
        & 46.1 & 48.0 & 52.7 & 1256 & 48.0 & 51.1 & 55.0 & 47.8 & 73.6 \\

        MustDrop \texttt{\scriptsize{(2024.11)}} 
        & 53.1 & 60.0 & 53.1 & 1612 & 68.0 & 63.4 & 69.3 & 54.2 & 88.5 \\

        PDrop \texttt{\scriptsize{(2024.10)}} 
        & 41.9 & 33.3 & 50.5 & 1092 & 55.9 & 68.6 & 69.2 & 45.9 & 74.5 \\

        VisionZip \texttt{\scriptsize{(CVPR25)}} 
        & 55.1 & 60.1 & 55.4 & 1690 & 77.0 & 69.0 & 72.4 & \textbf{55.5} & 93.1 \\

        SparseVLM \texttt{\scriptsize{(ICML25)}} 
        & 52.7 & 56.2 & 46.1 & 1505 & 75.1 & 62.2 & 68.2 & 51.8 & 85.6 \\

        HoloV \texttt{\scriptsize{(NeurIPS25)}} 
        & 55.3 & \textbf{63.3} & 55.1 & 1715 & 80.3 & 69.5 & 72.8 & 55.4 & 94.5 \\

        ApET \texttt{\scriptsize{(2026.02)}} 
        & 56.9 & 61.2 & 54.4 & 1714 & \textbf{84.4} & 68.9 & \textbf{75.1} & 53.9 & 94.6 \\

        \rowcolor{mygreen2}
        \textbf{\method{} (Ours)} 
        & \textbf{57.4} & 62.8 & \textbf{56.9} & \textbf{1754} & 83.8 & \textbf{70.0} & 75.0 & \textbf{55.5} & \textbf{96.4} \\
        
        \midrule

        % ================= Retain 32 ===================
        \rowcolor{mygray}
        LLaVA-1.5-7B & \multicolumn{8}{c}{\textit{Retain 32 Tokens} \ $\fg{(\text{Pruning Ratio}= 94.4\%)}$}\\

        ToMe \texttt{\scriptsize{(ICLR23)}} 
        & 43.6 & 31.6 & 28.1 & 828 & 39.0 & 41.4 & 46.8 & 38.3 & 55.3 \\

        FastV \texttt{\scriptsize{(ECCV24)}} 
        & 41.5 & 37.8 & 33.2 & 885 & 32.5 & 42.6 & 43.4 & 42.5 & 57.2 \\

        SparseVLM \texttt{\scriptsize{(ICML25)}} 
        & 48.3 & 51.4 & 40.6 & 1047 & 67.9 & 57.3 & 58.6 & 46.1 & 74.9 \\

        PruMerge+ \texttt{\scriptsize{(2024.05)}} 
        & 51.1 & 56.8 & 47.0 & 941 & 70.9 & 68.5 & 54.9 & 50.6 & 80.0 \\

        VisionZip \texttt{\scriptsize{(CVPR25)}} 
        & 51.8 & 57.7 & 50.3 & 1247 & 68.7 & 68.8 & 67.1 & 53.1 & 85.3 \\

        VisPruner \texttt{\scriptsize{(ICCV25)}} 
        & 52.2 & 58.4 & 52.7 & 1271 & 72.7 & 69.2 & 67.7 & 53.9 & 87.1 \\

        \rowcolor{mygreen2}
        \textbf{\method{} (Ours)} 
        & \textbf{55.1} & \textbf{61.2} & \textbf{55.3} & \textbf{1363} & \textbf{78.5} & \textbf{69.9} & \textbf{71.4} & \textbf{54.8} & \textbf{91.2} \\
        \bottomrule
	\end{tabular}
    }
    \caption{Comprehensive experimental results on LLaVA-1.5-7B across different image-language benchmarks under different pruning ratios. Best results are in \textbf{bold}. \rev{Unlike Tab.~\ref{tab:llava1-5}, this table additionally reports VQA-v2, so the averages are computed over eight benchmarks instead of seven and may differ slightly from those in the main paper.}}
    \label{tab:llava1-5-appendix}
\end{table*}
% \definecolor{mygreen2}{RGB}{238, 243, 243}
\begin{table*}[t]
    \centering
    \scriptsize
    \setlength{\tabcolsep}{4pt} 
    \begin{tabular}{l|cccc|c}
        \toprule
        \textbf{Method} 
        & \makecell{\textbf{NextQA}\\WUPS}
        & \makecell{\textbf{MVBench}\\Acc}
        & \makecell{\textbf{SEEDBench}\\Acc}
        & \makecell{\textbf{VideoMME}\\Score}
        & \makecell{\textbf{Average}\\(\%)} \\
        \midrule

        \rowcolor{gray!10}
        Qwen2.5-VL-7B (Full) & 26.22  & 68.10 & 62.18 & 60.67 & 100.0\% \\

        % FastV (ECCV24) & 25.67  & 65.75 & \textit{oom} & \textbf{59.15} & 97.3\% \\
        DART (EMNLP25)  & 25.52 & 65.80 & 61.00 & 57.74 & 96.8\% \\
        DivPrune (CVPR25) & 25.67 & 65.85 & 59.79 & 57.78 & 96.5\% \\

        \rowcolor{mygreen2}
        \textbf{\method{} (Ours)} & \textbf{26.03} & \textbf{66.70} & \textbf{61.80} & \textbf{58.59} & \textbf{98.3\%} \\

        \bottomrule
    \end{tabular}
    \caption{Comprehensive experimental results of various methods across different video-language benchmarks under an 80\% token pruning ratio. \rev{Unlike Tab.~\ref{tab:video_qwen25}, this table additionally reports NextQA, so the averages are computed over four benchmarks instead of three and may differ slightly from those in the main paper.}}
    \label{tab:video_appendix}
\end{table*}
\begin{table*}[t]
    \centering
    \tiny
    \setlength{\tabcolsep}{2.5pt}
    \resizebox{0.92\linewidth}{!}{
    \begin{tabular}{l|cccc|cccc}
        \toprule
        \textbf{Methods} 
        & \makecell{\textbf{Time}\\\textbf{(mm:ss)}}
        & \makecell{\textbf{Prefill}\\\textbf{(ms)}}
        & \makecell{\textbf{Latency}\\\textbf{(ms)}}
        & \makecell{\textbf{Acc.}\\\textbf{(\%)}}
        & \makecell{\textbf{Time}\\\textbf{(mm:ss)}}
        & \makecell{\textbf{Prefill}\\\textbf{(ms)}}
        & \makecell{\textbf{Latency}\\\textbf{(ms)}}
        & \makecell{\textbf{Acc.}\\\textbf{(\%)}} \\
        \midrule
        
        \textcolor{gray}{Upper Bound (576 Tokens)}
        & \textcolor{gray}{19:28}
        & \textcolor{gray}{62.75}
        & \textcolor{gray}{121.0}
        & \textcolor{gray}{100\%}
        & \textcolor{gray}{19:28}
        & \textcolor{gray}{62.75}
        & \textcolor{gray}{121.0}
        & \textcolor{gray}{100\%} \\
        \midrule
        
        \rowcolor{mygray}
        LLaVA-1.5-7B
        & \multicolumn{4}{c|}{\textit{Pruning Ratio = $\fg{\downarrow 77.8\%}$}}
        & \multicolumn{4}{c}{\textit{Pruning Ratio = $\fg{\downarrow 90.0\%}$}} \\

        FastV \texttt{\scriptsize{(ECCV24)}}
        & 13:56 & 42.75 & 83.6 & 69.4\%
        & 12:02 & 32.2 & 75.0 & 65.5\% \\
        
        VisionZip \texttt{\scriptsize{(CVPR25)}}
        & 13:28 & 44.11 & 85.4 & 95.1\%
        & 12:24 & 36.9 & 77.4 & 91.8\% \\
        
        SparseVLM \texttt{\scriptsize{(ICML25)}}
        & 13:07 & 46.01 & 83.5 & 94.6\%
        & 12:20 & 38.6 & 77.0 & 90.5\% \\
        
        \rowcolor{mygreen2}
        \textbf{\method{} \scriptsize{(Ours)}}
        & 14:01 & 45.24 & 89.8 & 99.8\%
        & 12:59 & 37.1 & 86.0 & 97.1\% \\
        \bottomrule
    \end{tabular}
    }
    \caption{Real inference comparison on POPE. Experiments adopt 77.8\% and 90\% pruning ratios.}
    \label{tab:efficiency_appendix}
\end{table*}

\section{Generalization across Model Scales and Visual-Token Regimes}
\label{app:generalization}

\rev{The main paper evaluates fixed-grid LLaVA-1.5-7B and dynamic-resolution Qwen2.5-VL-7B. This section reports two further settings that stress the two remaining axes of variation: model capacity and visual-sequence length. In both cases the vision encoder, the top-$\rho$ rule, and all other hyperparameters are kept unchanged.}

\subsection{Transfer across Model Scale}
\label{app:scale_13b}

\rev{We first run \textit{\method{}} on LLaVA-1.5-13B, replacing only the language model. As reported in Tab.~\ref{tab:llava_13b}, the behavior is consistent with the 7B setting. With 192 retained tokens, \textit{\method{}} preserves 99.2\% of the full-model average and even slightly exceeds the unpruned model on MME and SQA. Under roughly 90\% pruning (64 tokens), it preserves 96.2\% and outperforms VisionZip by 2.0 relative-performance points, with the largest single-benchmark gap on POPE (84.77 vs.\ 76.00). This indicates that high-norm outliers are a property of the visual representation rather than of one specific decoder capacity.}

% =============================================================================
% NEW TABLE for the camera-ready version.
% Source: rebuttal to Reviewer rBoA, Q1a (Generalization across model scale).
% =============================================================================
\begin{table}[t]
    \centering
    \setlength{\tabcolsep}{3.0pt}
    \renewcommand{\arraystretch}{1.1}
    \scriptsize

    \begin{tabular}{l | c | *{4}{>{\centering\arraybackslash}p{0.82cm}} | c}
    \toprule
        \textbf{Methods} & \textbf{Tok} & \textbf{GQA} & \textbf{MME} & \textbf{POPE} & \textbf{SQA} & \textbf{Avg}\\
        \midrule

        \textcolor{gray}{Upper Bound}
        & \textcolor{gray}{576}
        & \textcolor{gray}{63.25}
        & \textcolor{gray}{1827.05}
        & \textcolor{gray}{87.13}
        & \textcolor{gray}{72.73}
        & \textcolor{gray}{100.0\%} \\
        \midrule

        VisionZip \texttt{\scriptsize{(CVPR25)}}
        & 192 & 59.60 & 1770.00 & 86.40 & 72.80 & 97.8\% \\

        \rowcolor{mygreen2}
        \textbf{\method{} (Ours)}
        & 192 & \textbf{60.04} & \textbf{1839.36} & \textbf{87.11} & \textbf{73.53} & \textbf{99.2\%} \\
        \midrule

        VisionZip \texttt{\scriptsize{(CVPR25)}}
        & 64 & 56.20 & 1676.00 & 76.00 & \textbf{74.40} & 94.2\% \\

        \rowcolor{mygreen2}
        \textbf{\method{} (Ours)}
        & 64 & \textbf{57.60} & \textbf{1750.13} & \textbf{84.77} & 73.33 & \textbf{96.2\%} \\
        \bottomrule
    \end{tabular}

    \caption{\rev{Transfer across model scale on LLaVA-1.5-13B. \textit{\method{}} keeps its advantage when the language model grows from 7B to 13B. Best results per budget are in \textbf{bold}.}}
    \label{tab:llava_13b}
\end{table}

\subsection{Transfer to High-Resolution Inputs}
\label{app:highres_next}

\rev{We next evaluate on LLaVA-NeXT-7B~\cite{Llava_next}, whose any-resolution scheme expands one image into roughly $2{,}880$ visual tokens---five times the $576$ tokens of LLaVA-1.5---placing the high-norm observation in a markedly different token regime. All numbers in Tab.~\ref{tab:llava_next} are produced by a single end-to-end run of the same evaluation harness, and the \textbf{Avg} column normalizes every row against the Upper Bound measured in that run.}

\rev{\textit{\method{}} attains the best average retention at both budgets, reaching 94.2\% with 320 retained tokens and 92.2\% with 160, ahead of the strongest baseline VisPruner by 2.0 and 5.7 points. The margin widens as the budget tightens, which mirrors the trend we observe on LLaVA-1.5: the harder the compression, the more it matters that the retained tokens are not high-norm background. The advantage is concentrated on POPE, where \textit{\method{}} loses almost nothing relative to the full model (87.40 vs.\ 87.61, \ie, 99.8\% retention) while every baseline degrades by at least 7 points, and on GQA, where it leads VisPruner by 1.8 and 3.9 points.}

\rev{TextVQA is the one benchmark on which \textit{\method{}} does not lead. It retains 89.0\% (320 tokens) and 88.8\% (160 tokens) of the full-model score, close to VisionZip (90.1\% and 89.1\%) but below VisPruner (93.9\% and 91.2\%). We attribute the residual gap to the interaction between the sanitizer and scene text under tiled high-resolution encoding: small glyph patches can themselves carry large feature norms, so a purely norm-based rule may aggregate a few genuine textual tokens together with background outliers. Notably, the gap does not widen at the tighter budget (2.98 points at 320 tokens versus 1.51 at 160), indicating that the effect is a bounded bias of the criterion rather than a breakdown under aggressive pruning; on LLaVA-1.5 and Qwen2.5-VL, where visual tokens are not tiled, \textit{\method{}} gives the best TextVQA scores among all compared methods. Making the sanitizer text-aware, for instance by exempting high-norm tokens with high local edge density, is left to future work.}

% =============================================================================
% NEW TABLE for the camera-ready version.
% Source: rebuttal to Reviewer rBoA, Q1b (Generalization to high-resolution
% architecture). LLaVA-NeXT-7B produces ~2880 visual tokens.
%
% UPDATED with the end-to-end 8xH800 re-run. Two things changed w.r.t. the
% rebuttal version:
%   (1) the Upper Bound TextVQA score is 61.37 (previously 64.80), and
%   (2) SinkPruner's TextVQA scores improved markedly (54.62 / 54.49).
% Because the Upper Bound is the denominator of the normalized Avg column, the
% Avg of EVERY row (baselines included) has been recomputed against the new
% Upper Bound so that all rows share the same denominator:
%     Avg = mean(GQA/64.23, TextVQA/61.37, POPE/87.61) x 100
% Baseline raw scores are unchanged.
% =============================================================================
\definecolor{mygreen2}{RGB}{238, 243, 243}

\begin{table}[t]
    \centering
    \setlength{\tabcolsep}{3.0pt}
    \scriptsize

    \begin{tabular}{l | *{3}{>{\centering\arraybackslash}p{1.0cm}} |>{\centering\arraybackslash}p{0.85cm}}
    \toprule
        \textbf{Methods} & \textbf{GQA} & \textbf{VQA}$_{\text{Text}}$ & \textbf{POPE} & \textbf{Avg}\\
        \midrule

        \textcolor{gray}{Upper Bound, 2880 Tokens}
        & \textcolor{gray}{64.23}
        & \textcolor{gray}{61.37}
        & \textcolor{gray}{87.61}
        & \textcolor{gray}{100.0\%} \\
        \midrule

        % ================= Retain 320 ===================
        \rowcolor{mygray}
        LLaVA-NeXT-7B & \multicolumn{4}{c}{\textit{Retain 320 Tokens} \ $\fg{(\text{Pruning Ratio}\approx 88.9\%)}$}\\

        FastV \texttt{\scriptsize{(ECCV24)}}
        & 55.90 & 55.70 & 71.70 & 86.5 \\

        SparseVLM \texttt{\scriptsize{(ICML25)}}
        & 56.50 & 52.40 & 73.50 & 85.7 \\

        VisionZip \texttt{\scriptsize{(CVPR25)}}
        & 58.10 & 55.30 & 75.00 & 88.7 \\

        VisPruner \texttt{\scriptsize{(ICCV25)}}
        & 58.40 & \textbf{57.60} & 80.40 & 92.2 \\

        \rowcolor{mygreen2}
        \textbf{\method{} (Ours)}
        & \textbf{60.19} & 54.62 & \textbf{87.40} & \textbf{94.2} \\

        \midrule

        % ================= Retain 160 ===================
        \rowcolor{mygray}
        LLaVA-NeXT-7B & \multicolumn{4}{c}{\textit{Retain 160 Tokens} \ $\fg{(\text{Pruning Ratio}\approx 94.4\%)}$}\\

        FastV \texttt{\scriptsize{(ECCV24)}}
        & 49.80 & 51.90 & 51.70 & 73.7 \\

        SparseVLM \texttt{\scriptsize{(ICML25)}}
        & 50.20 & 45.10 & 54.60 & 71.3 \\

        VisionZip \texttt{\scriptsize{(CVPR25)}}
        & 54.30 & 54.70 & 59.40 & 80.5 \\

        VisPruner \texttt{\scriptsize{(ICCV25)}}
        & 54.70 & \textbf{56.00} & 72.90 & 86.5 \\

        \rowcolor{mygreen2}
        \textbf{\method{} (Ours)}
        & \textbf{58.56} & 54.49 & \textbf{84.67} & \textbf{92.2} \\
        \bottomrule
    \end{tabular}
    \caption{\rev{Performance comparison on the high-resolution LLaVA-NeXT-7B, whose visual input expands to about $2{,}880$ tokens. \textbf{Avg} is the mean per-benchmark ratio to the Upper Bound. Best results are in \textbf{bold}.}}
    \label{tab:llava_next}
    \vspace{-10pt}
\end{table}

\section{Ablations of the Remaining Hyperparameters}
\label{app:hyper_ablation}

\rev{Beyond the high-norm criterion analyzed in Sec.~\ref{app:sensitivity_analysis}, \textit{\method{}} involves two further groups of design choices: \emph{where} the text-guided pruner acts inside the LLM together with its progressive retention schedule, and \emph{how} the low-norm budget is split between the salience pool and the diversity pool. We stress that these are empirical implementation defaults rather than the core contribution, and that we did not perform a grid search to select them. The analyses below verify that claim.}

\smallskip
\noindent\rev{\textbf{Pruning layers and progressive schedules.}}
\rev{Our default prunes at layers $(2, 6, 15)$, which realizes a coarse-to-fine rule: early layers discard obvious redundancy, middle layers remove cross-modally irrelevant tokens, and later layers retain text-relevant evidence once cross-modal interaction has become discriminative. To test whether performance hinges on this exact triplet, we perturb each pruning layer by $\pm 1$ while keeping the layer-wise token-computation budget $\sum_l n_l$ identical across configurations; the retention schedule is adjusted accordingly so that all seven settings consume the same compute.
Tab.~\ref{tab:layer_sensitivity} shows a flat response surface. POPE accuracy stays within a $1.21$-point band ($80.17$ to $81.38$) and MME Total within about $35$ points, and no configuration collapses. Notably, our reported setting is \emph{not} the best-performing one, which is consistent with our statement that it was never optimized per benchmark. We note that this analysis establishes insensitivity to pruning \emph{locations} and retention allocation; it does not claim that three stages are theoretically optimal, and we adopt three stages simply as a progressive default that balances gradual pruning against implementation overhead.}

\smallskip
\noindent\rev{\textbf{Salience-to-diversity split.}}
\rev{The visual sanitizer draws representative tokens from a salience pool and a diversity pool. To separate the effect of the \emph{split} from that of the \emph{budget}, we fix the total candidate budget to $2\times$Retain and vary only the allocation between the two pools.
As shown in Tab.~\ref{tab:pool_split}, POPE accuracy spans $80.49$ to $81.91$ and MME Total spans $1629.05$ to $1682.61$, with the two metrics preferring opposite ends of the range: diversity-heavy splits favor POPE while salience-heavy splits favor MME. Rather than picking whichever endpoint maximizes a particular benchmark, we keep the symmetric $5\!:\!5$ split as a benchmark-independent default.}

% =============================================================================
% NEW TABLES for the camera-ready version.
% Source: rebuttal to Reviewer rBoA, Q3 (Hyperparameter ablations) and
%         rebuttal to Reviewer ZYCL, Q2 (Stage-2 architectural choices).
% Both studies are run on LLaVA-1.5-7B under the Retain-32 configuration.
% =============================================================================

\begin{table*}[t]
    \centering
    \setlength{\tabcolsep}{6pt}
    \renewcommand{\arraystretch}{1.15}
    \small

    \begin{tabular}{l c c cc ccc}
    \toprule
        \textbf{Config} & \textbf{Prune layers} & \textbf{Retain schedule}
        & \textbf{POPE Acc.} & \textbf{POPE F1}
        & \textbf{MME-P} & \textbf{MME-C} & \textbf{MME Total} \\
        \midrule

        \rowcolor{mygreen2}
        Default (ours)          & $[2, 6, 15]$ & $[54, 36, 22]$ & 80.93 & 77.52 & 1333.84 & 296.79 & 1630.63 \\
        A1: $2 \!\to\! 1$       & $[1, 6, 15]$ & $[56, 36, 22]$ & 80.69 & 77.13 & 1356.92 & 299.64 & 1656.56 \\
        A2: $2 \!\to\! 3$       & $[3, 6, 15]$ & $[51, 36, 22]$ & 81.09 & 77.71 & 1332.25 & 301.43 & 1633.68 \\
        B1: $6 \!\to\! 5$       & $[2, 5, 15]$ & $[54, 38, 22]$ & 80.17 & 76.20 & 1333.10 & 303.93 & 1637.02 \\
        B2: $6 \!\to\! 7$       & $[2, 7, 15]$ & $[54, 34, 22]$ & \textbf{81.38} & \textbf{78.22} & \textbf{1358.95} & \textbf{306.79} & \textbf{1665.74} \\
        C1: $15 \!\to\! 14$     & $[2, 6, 14]$ & $[54, 36, 23]$ & 81.01 & 77.64 & 1348.86 & 296.43 & 1645.29 \\
        C2: $15 \!\to\! 16$     & $[2, 6, 16]$ & $[54, 36, 21]$ & 80.90 & 77.47 & 1339.32 & 306.43 & 1645.75 \\
        \bottomrule
    \end{tabular}

    \caption{\rev{Sensitivity to the choice of pruning layers. Each pruning layer is perturbed by $\pm1$ while the layer-wise token-computation budget $\sum_l n_l$ is held constant across configurations, so that only the \emph{location} of pruning changes. POPE accuracy varies within a $1.21$-point band and MME Total within about $35$ points; the configuration used in the paper is not the best-performing one, confirming that it was not selected by benchmark-specific search.}}
    \label{tab:layer_sensitivity}
\end{table*}

\begin{table}[t]
    \centering
    \setlength{\tabcolsep}{4pt}
    \renewcommand{\arraystretch}{1.1}
    \scriptsize

    \begin{tabular}{c cc cc c}
    \toprule
        \textbf{Sal.\,:\,Div.} & \textbf{Sal.} & \textbf{Div.}
        & \makecell{\textbf{POPE}\\\textbf{Acc.}} & \makecell{\textbf{POPE}\\\textbf{F1}}
        & \makecell{\textbf{MME}\\\textbf{Total}} \\
        \midrule
        $3:7$ & 19 & 45 & 81.86 & \textbf{79.01} & 1629.05 \\
        $4:6$ & 26 & 38 & \textbf{81.91} & 78.93 & 1662.61 \\
        \rowcolor{mygreen2}
        $5:5$ (ours) & 32 & 32 & 80.93 & 77.52 & 1630.63 \\
        $6:4$ & 38 & 26 & 80.52 & 76.90 & 1675.48 \\
        $7:3$ & 45 & 19 & 80.49 & 76.72 & \textbf{1682.61} \\
        \bottomrule
    \end{tabular}

    \caption{\rev{Allocation between the salience pool and the diversity pool under a fixed total budget of $2\times$Retain, which isolates the effect of the split from that of the budget. The response surface is smooth, and we keep the symmetric $5\!:\!5$ split as a benchmark-independent default.}}
    \label{tab:pool_split}
\end{table}

\section{Additional Analysis of the Pruning Mechanism}
\label{app:mechanism_analysis}

In this section, we provide additional analysis to better understand why \textit{\method{}} consistently outperforms prior text-guided pruning methods such as SparseVLM.
Our analysis suggests that previous methods are affected by two closely related issues, namely \textit{massive activations} and \textit{text-visual attention dispersion}, both of which are substantially alleviated by our cascading design.

\subsection{Mitigating Massive Activations}
\label{app:analysis_massive_activation}

Existing text-guided methods (\eg, SparseVLM~\cite{SparseVLM25}) are built on the assumption that large attention weights in the LLM decoder indicate semantic relevance.
However, recent studies~\cite{visual_atten_sink25} show that MLLM decoders often exhibit \textit{massive activations}, where a small set of tokens attract disproportionately large attention regardless of their textual relevance.
These tokens behave as \textit{attention sinks}, which can distort token importance estimation and mislead downstream pruning.

\begin{figure}[t!]
    \centering
    \includegraphics[width=0.9\linewidth]{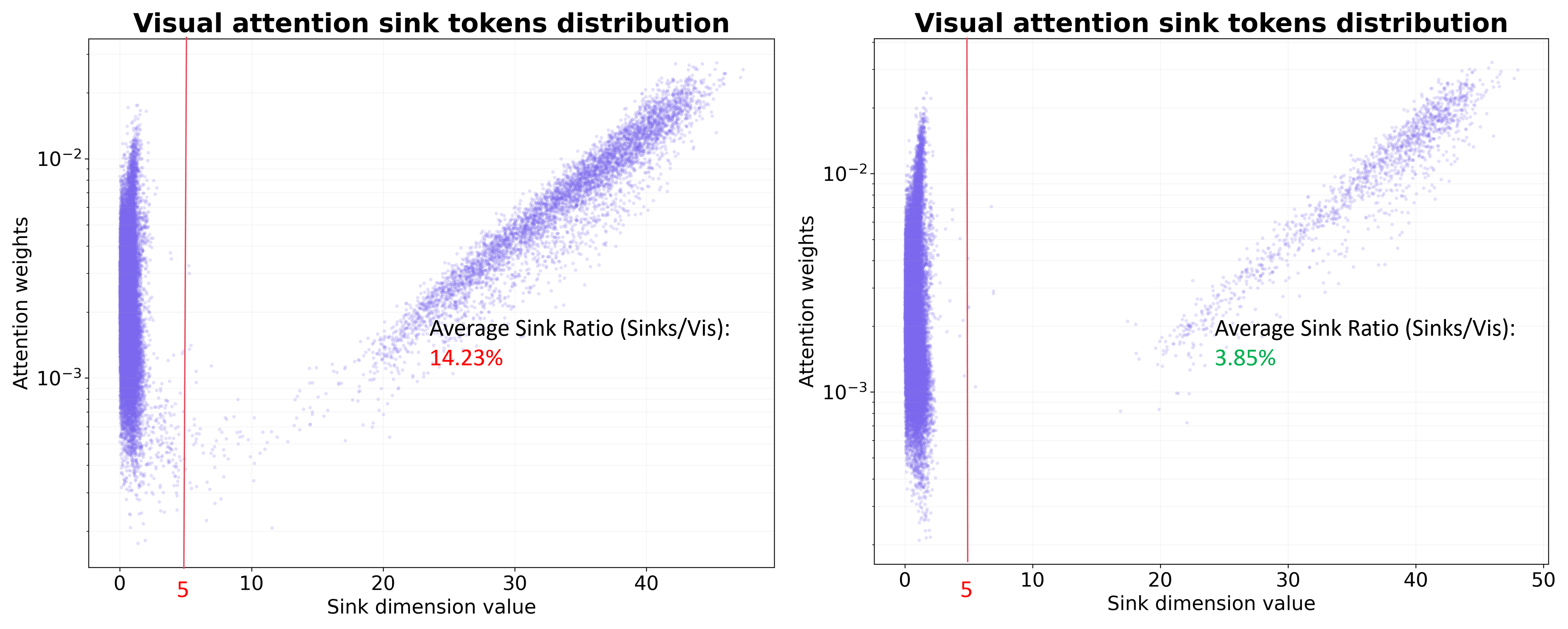}
    \caption{\textbf{Visual Attention Sinks in the LLM Decoder.} 
    The scatter plot correlates attention weights (y-axis) with massive activation values (x-axis). 
    \textbf{(Left)} Standard text-guided methods (\eg, SparseVLM) operate in a \textit{high-sink regime}, retaining a large cluster of attention sinks (red line, ratio 14.23\%).
    \textbf{(Right)} \textit{\method{}} operates in a \textit{low-sink regime}. By applying our \textit{visual sanitizer} to reduce inherent visual redundancy, we effectively suppress massive activations, reducing the sink ratio to 3.85\% and clarifying the attention landscape.}
    \label{fig:decoder_sink}
\end{figure}

Our empirical analysis shows that \textbf{reducing inherent visual redundancy significantly alleviates this effect}.
As shown in Fig.~\ref{fig:decoder_sink}, standard text-guided pruning operates on raw visual sequences and therefore tends to retain many sink tokens due to their inflated attention scores.
This leads to a sink ratio of \textbf{14.23\%}, meaning that a noticeable fraction of the token budget is consumed by non-semantic outliers.
In contrast, our \textit{visual sanitizer} first purifies the visual stream before it enters the LLM decoder.
This upstream pre-conditioning substantially suppresses the massive activation phenomenon, reducing the sink ratio to \textbf{3.85\%}.
As a result, the following text-guided pruning stage can operate on a much cleaner attention landscape and focus more reliably on truly relevant visual evidence.

\subsection{Reducing Text-Visual Attention Dispersion}
\label{app:analysis_entropy}

\begin{table}[t]
    \centering
    \small
    \setlength{\tabcolsep}{2.8pt}
    \caption{\textbf{Text-visual attention entropy at the pruning layer.} Lower entropy indicates higher selection confidence.}
    \label{tab:entropy}
    \begin{tabular}{@{}lc@{}}
    \toprule
    \textbf{Method} & \textbf{Entropy} ($\downarrow$) \\
    \midrule
    SparseVLM & 6.359 \\
    \textbf{\method{} (Ours)} & \textbf{4.851} \\
    \bottomrule
    \end{tabular}
    \vspace{-10pt}
\end{table}

A second limitation of text-guided pruning methods is \textit{text-visual attention dispersion}.
When raw and highly redundant visual tokens are directly fed into the LLM, cross-modal attention tends to become diffuse and uncertain.
We quantify this behavior using Shannon entropy: high entropy indicates that attention is broadly spread and the model lacks discriminative focus, whereas low entropy indicates a sharper and more confident selection pattern.

As shown in Tab.~\ref{tab:entropy}, SparseVLM exhibits a relatively high attention entropy of 6.36, reflecting substantial selection uncertainty caused by redundant visual inputs.
By contrast, \textit{\method{}} significantly reduces the entropy to \textbf{4.85}.
This result suggests that the \textit{visual sanitizer} acts as an effective denoising stage: by filtering redundant visual content before text-guided pruning, it sharpens the text-to-vision attention distribution and enables the model to identify linguistically relevant tokens with higher confidence.

\section{Additional Visualizations}
\label{app:more_visuals}

In this section, we provide more qualitative visualizations to better illustrate the observations behind our method.
These examples offer direct evidence that the problematic attention outliers discussed in the main paper are closely associated with high-norm tokens, and that such tokens are typically concentrated in spatially redundant, non-semantic regions.

\subsection{Visualizations of attention outliers and high-norm tokens}
\begin{figure}[t]
    \centering
    \includegraphics[width=0.7\linewidth]{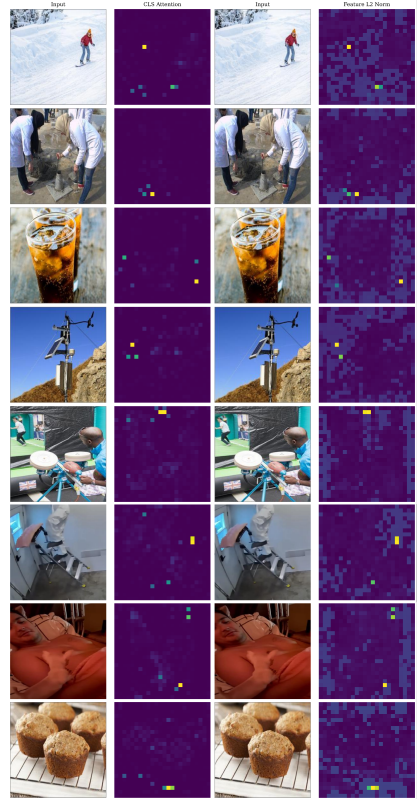}
    \caption{\textbf{Attention Artifacts Coincide with High-Norm Outliers.}
    We show multiple image examples (left to right): \emph{Input}, \emph{CLS Attention}, \emph{Input}, and \emph{Feature $\ell_2$ Norm}.
    The CLS attention maps exhibit sparse, peaky outliers (bright spots) that frequently appear in non-semantic background regions.
    Crucially, these attention peaks spatially align with abnormally large feature norms in the corresponding $\ell_2$-norm heatmaps, indicating that the observed high attention is largely driven by \textbf{high-norm outliers} rather than semantic relevance.}
    \label{fig:app_attn_norm_vis1}
\end{figure}

Figure~\ref{fig:app_attn_norm_vis1} presents multiple examples comparing \texttt{CLS} attention maps and feature $\ell_2$-norm heatmaps.
The results show that sparse attention spikes consistently align with a small set of abnormally high-norm tokens, confirming that these outliers are responsible for the distorted attention patterns observed in the vision encoder.
See Fig.~\ref{fig:app_attn_norm_vis1}.

\subsection{Visualizations of high-norm patches and low-norm patches}
\begin{figure*}[t]
\centering
\begin{minipage}[t]{0.38\textwidth}
    \centering
    \includegraphics[width=\linewidth]{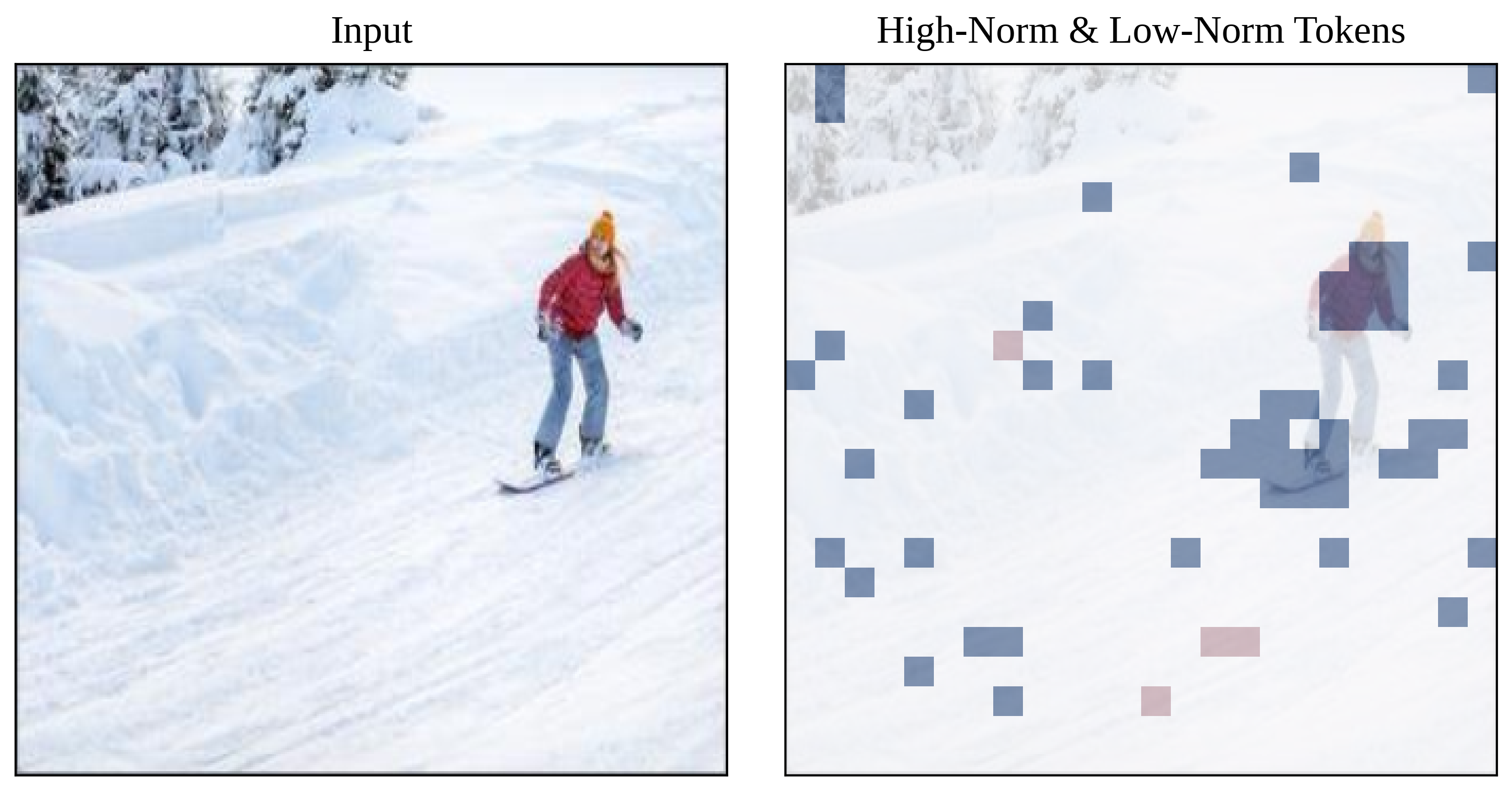}\par\vspace{2pt}
    \includegraphics[width=\linewidth]{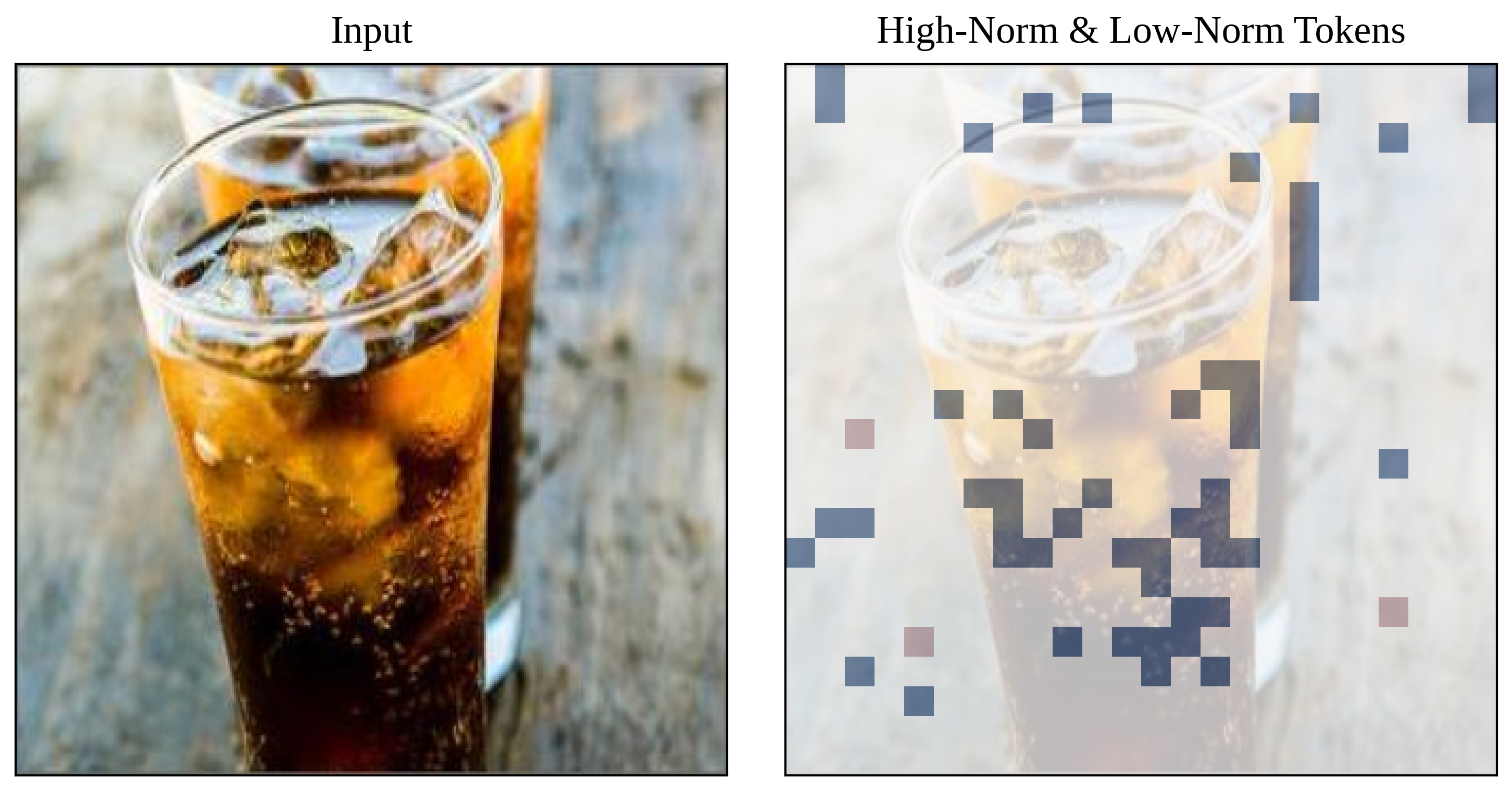}\par\vspace{2pt}
    \includegraphics[width=\linewidth]{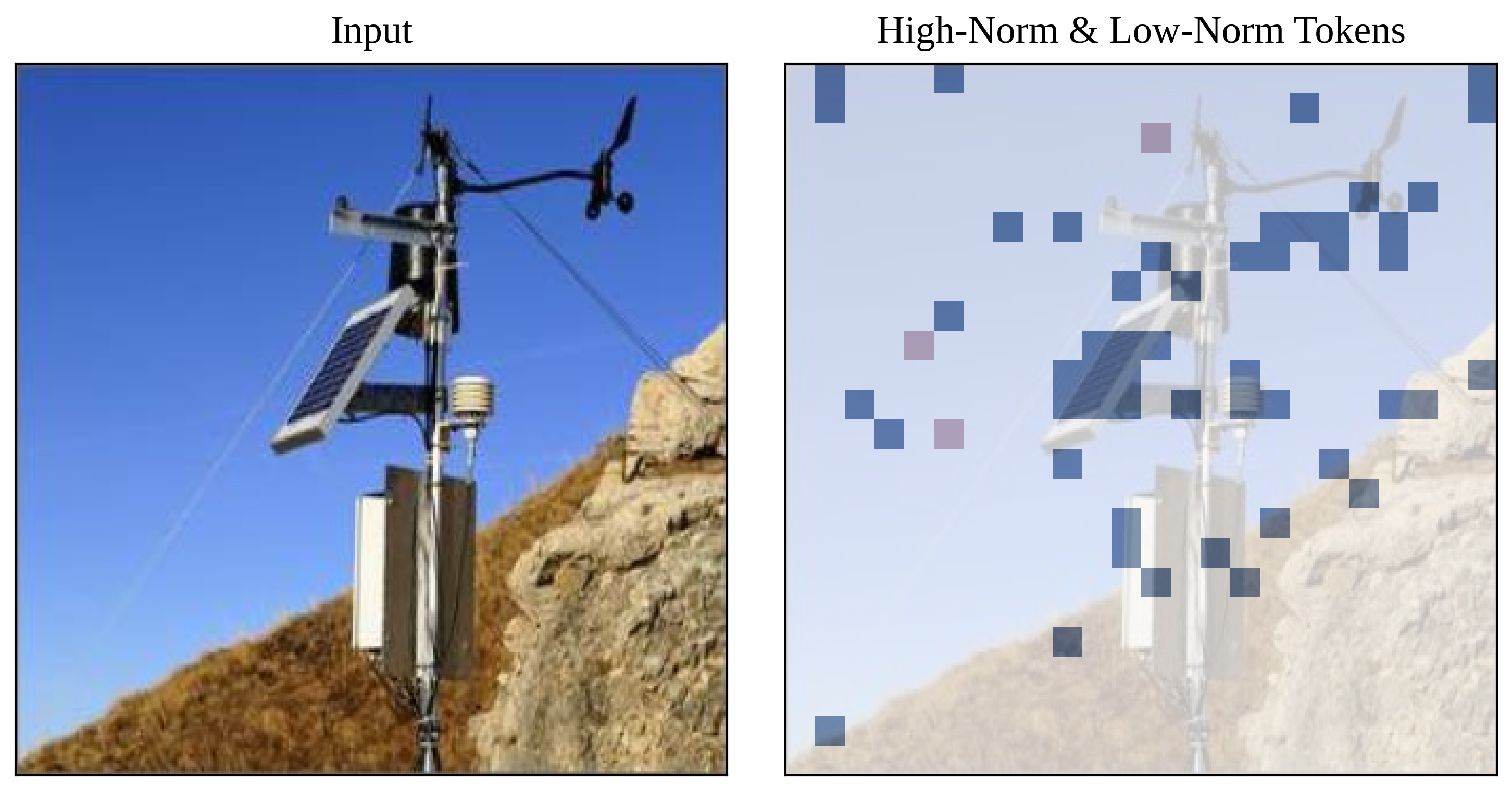}\par\vspace{2pt}
    \includegraphics[width=\linewidth]{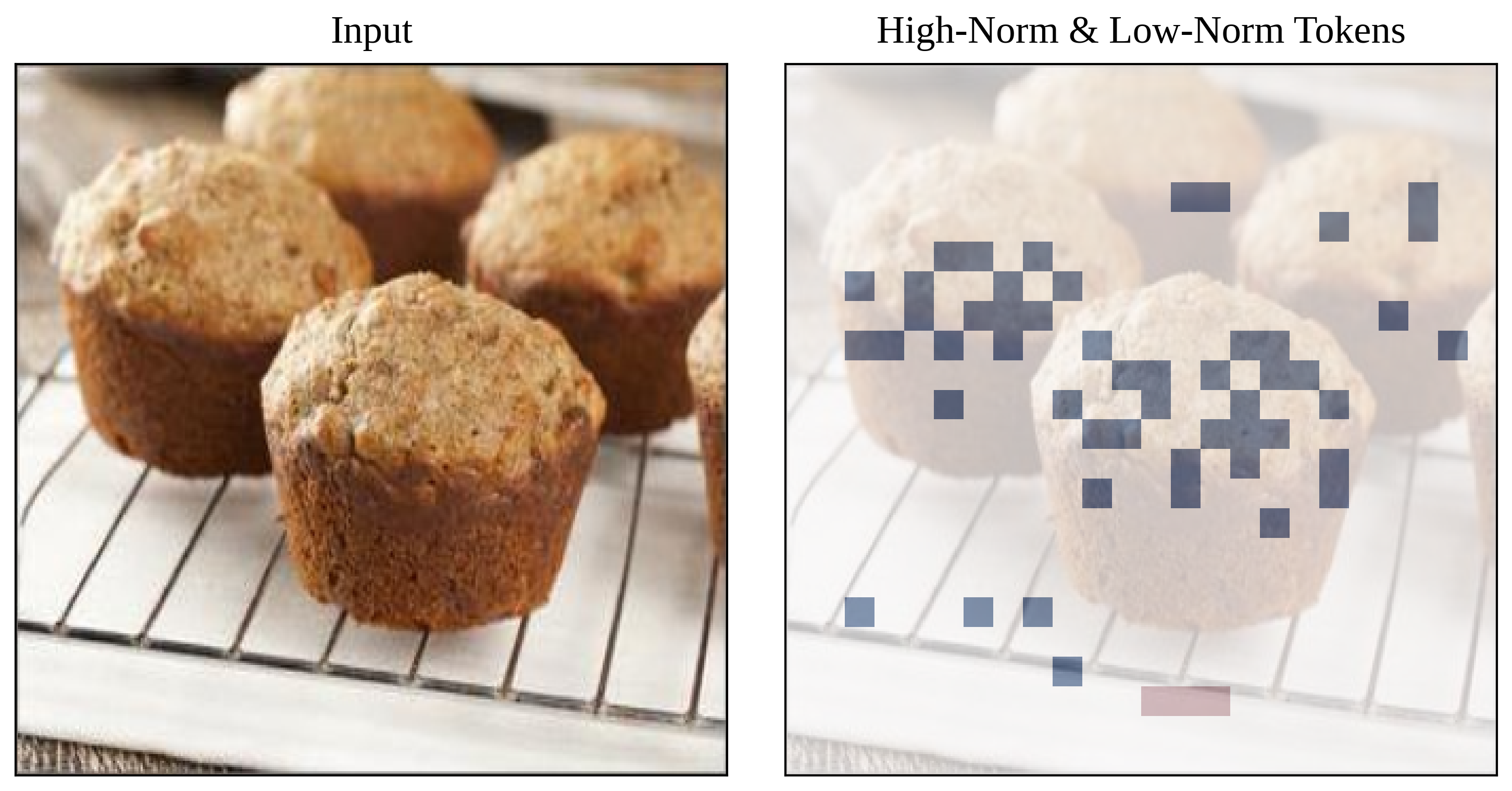}\par\vspace{2pt}
    \includegraphics[width=\linewidth]{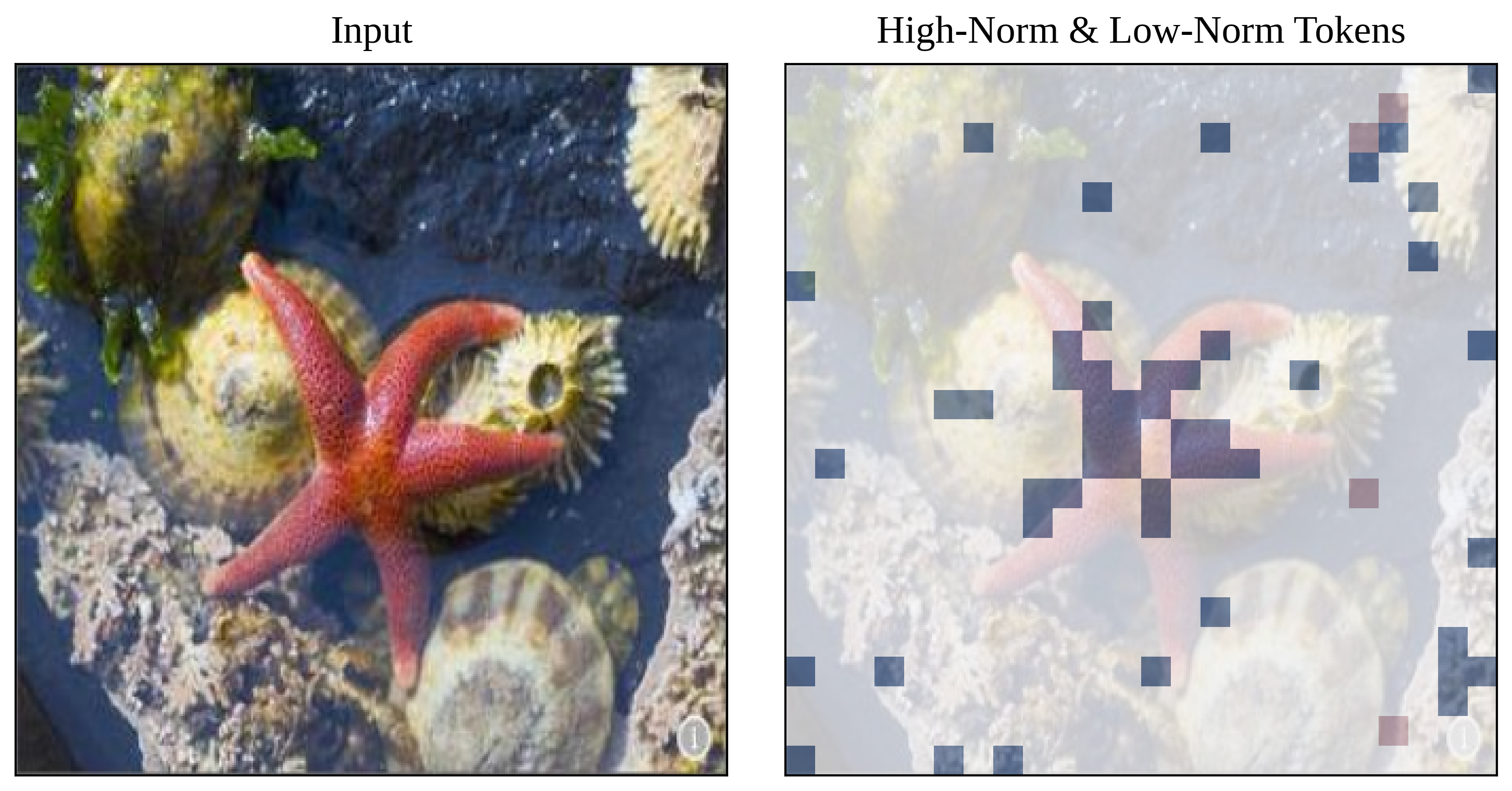}\par\vspace{2pt}
    \includegraphics[width=\linewidth]{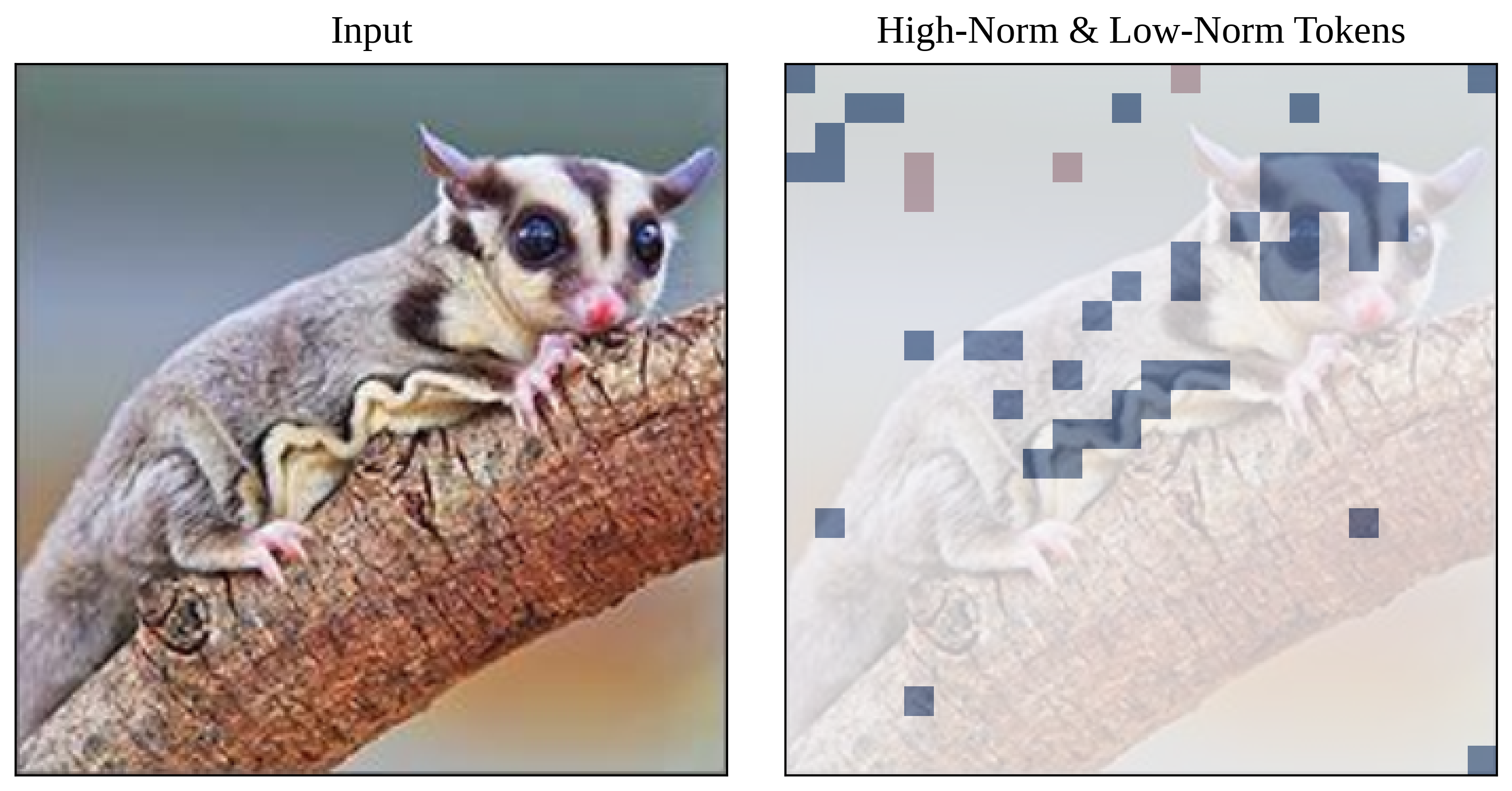}\par\vspace{2pt}
    \includegraphics[width=\linewidth]{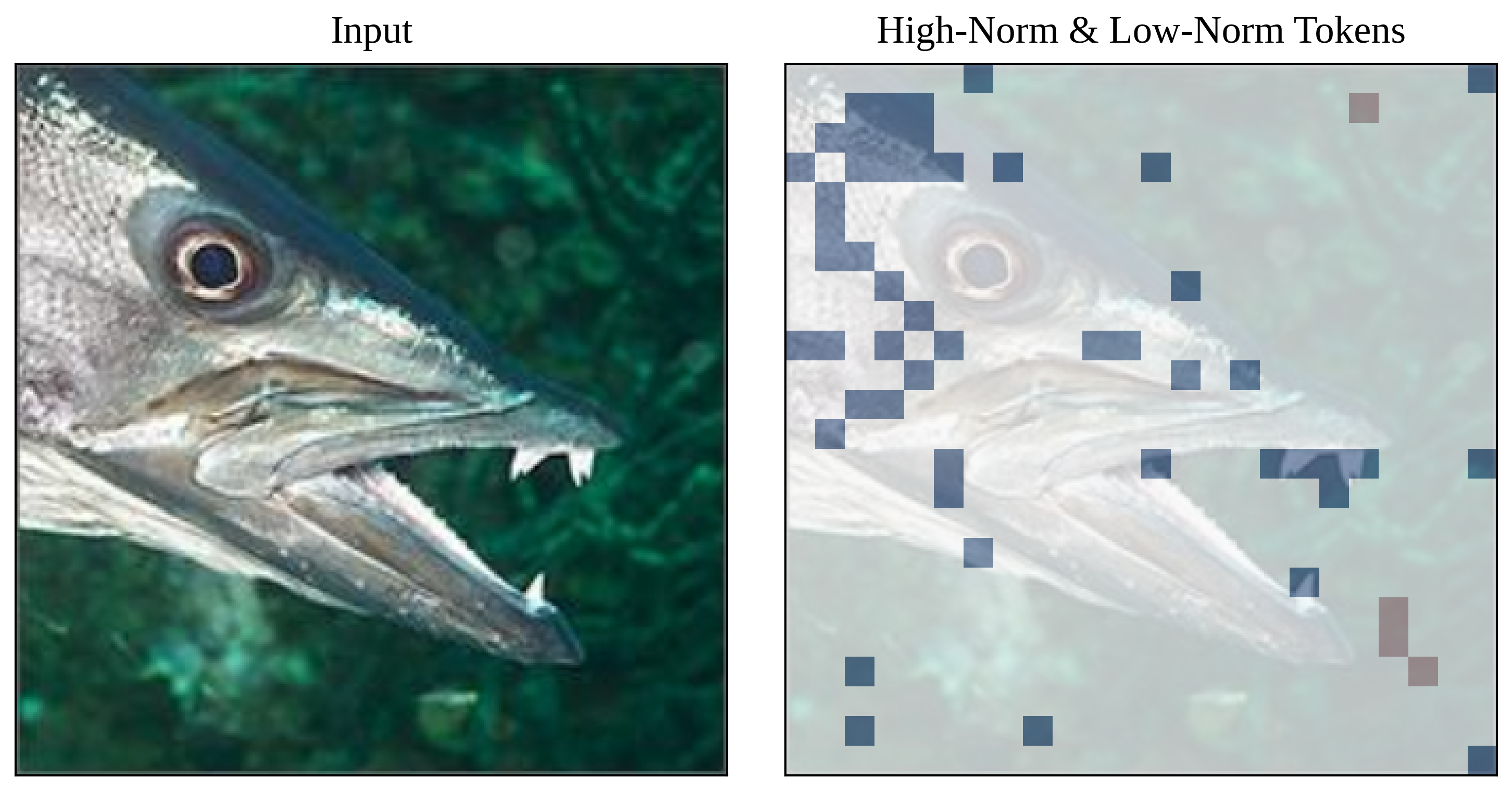}
\end{minipage}
\hspace{0.003\textwidth}
\begin{minipage}[t]{0.38\textwidth}
    \centering
    \includegraphics[width=\linewidth]{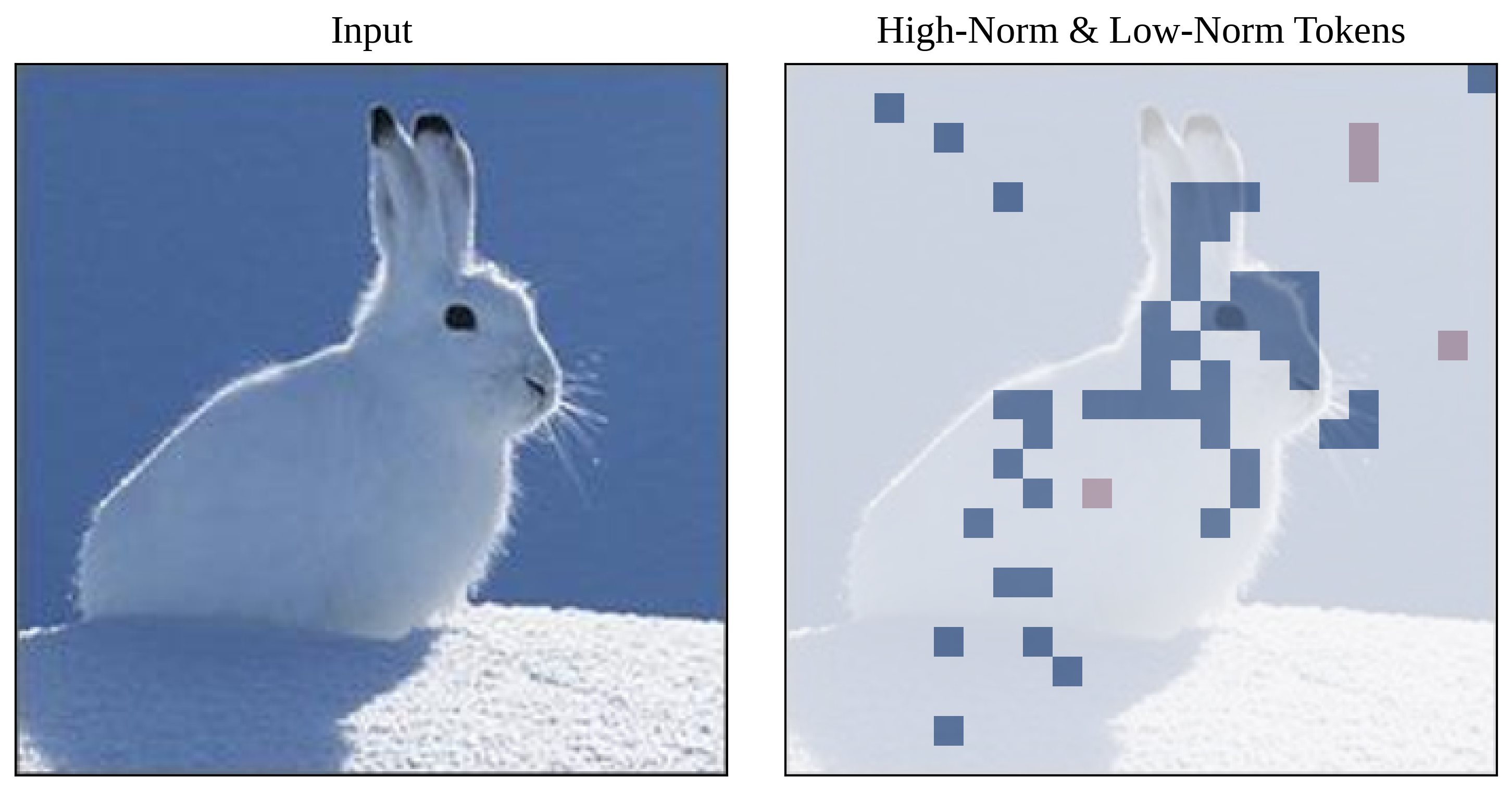}\par\vspace{2pt}
    \includegraphics[width=\linewidth]{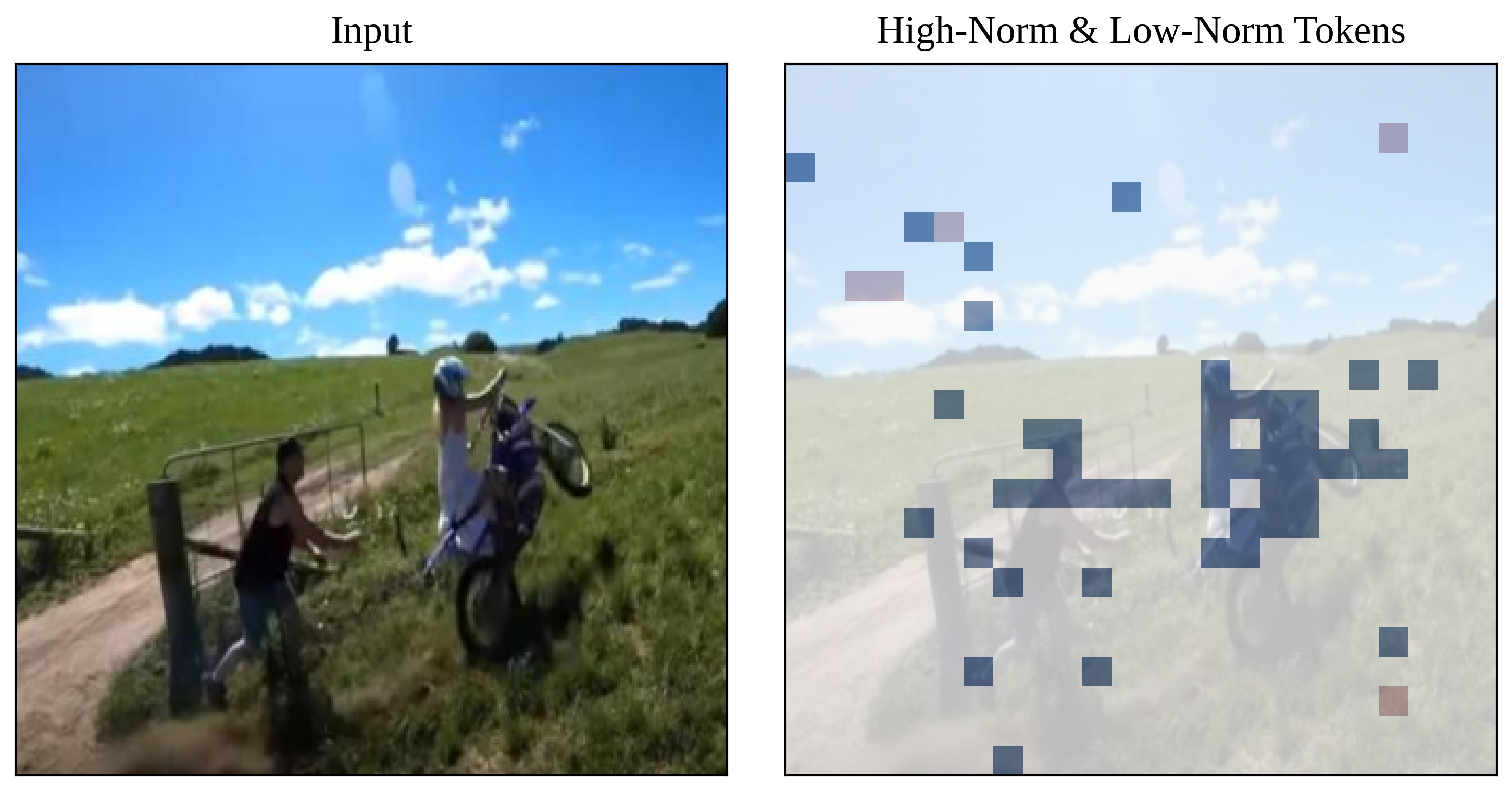}\par\vspace{2pt}
    \includegraphics[width=\linewidth]{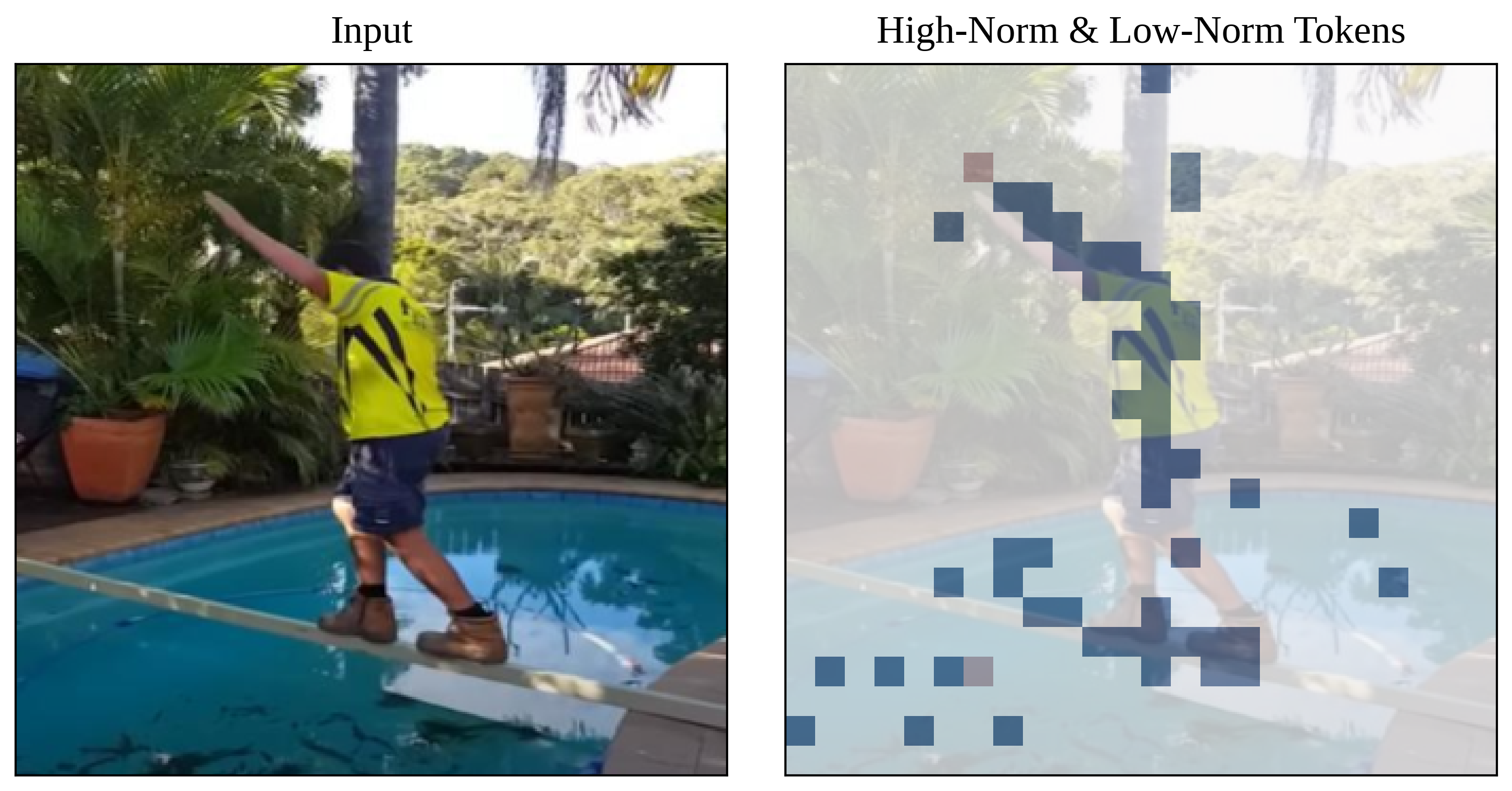}\par\vspace{2pt}
    \includegraphics[width=\linewidth]{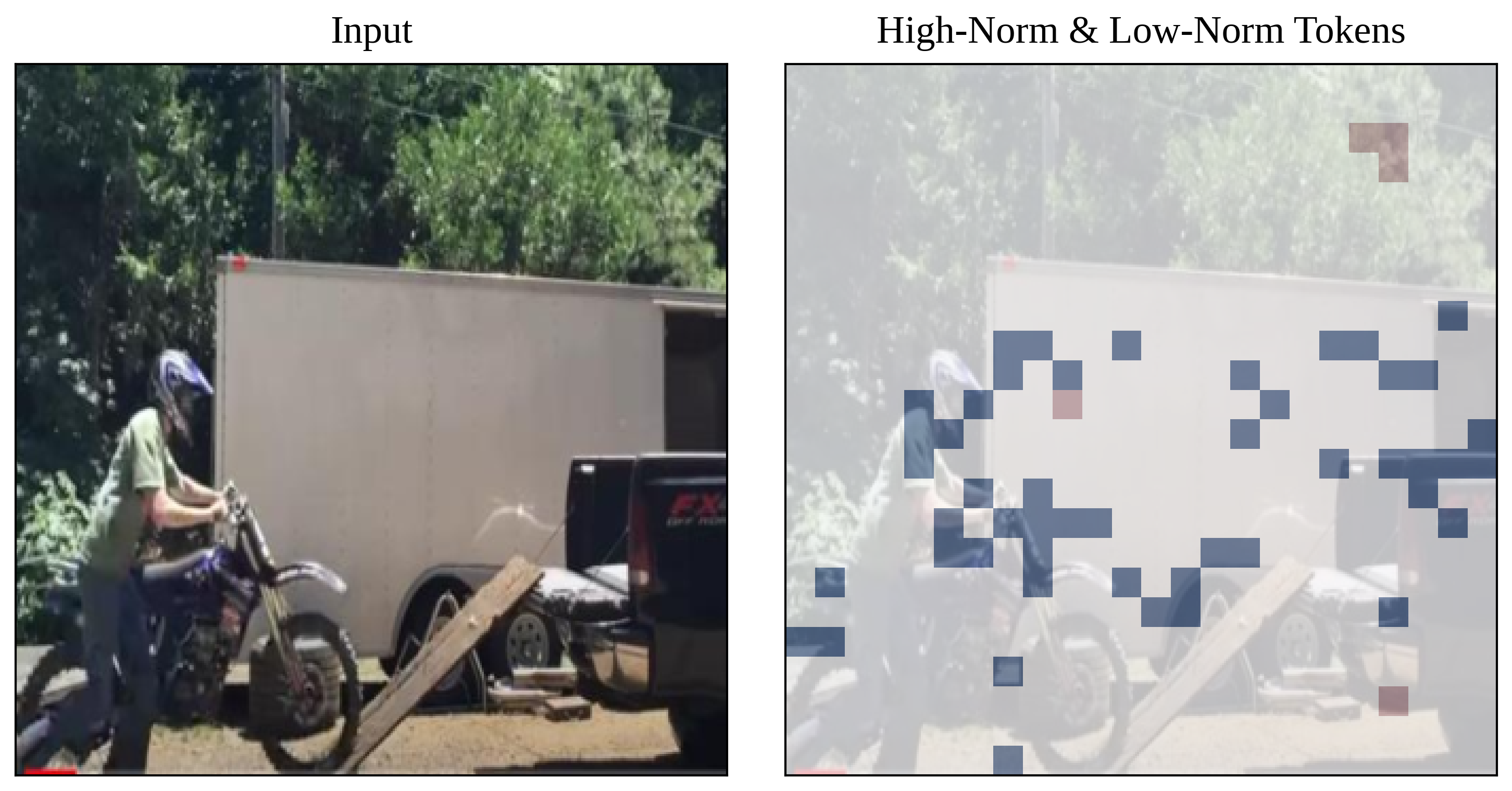}\par\vspace{2pt}
    \includegraphics[width=\linewidth]{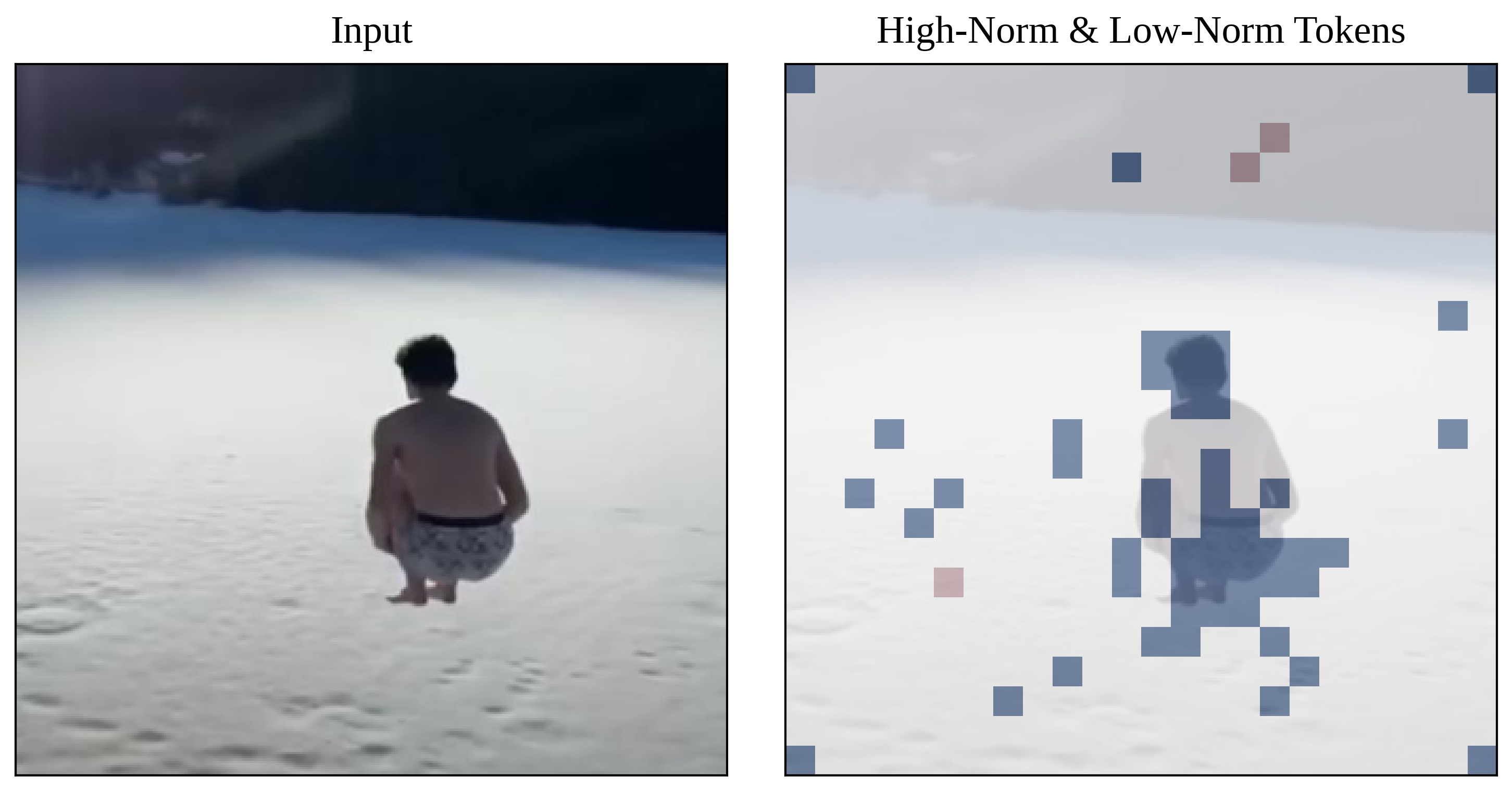}\par\vspace{2pt}
    \includegraphics[width=\linewidth]{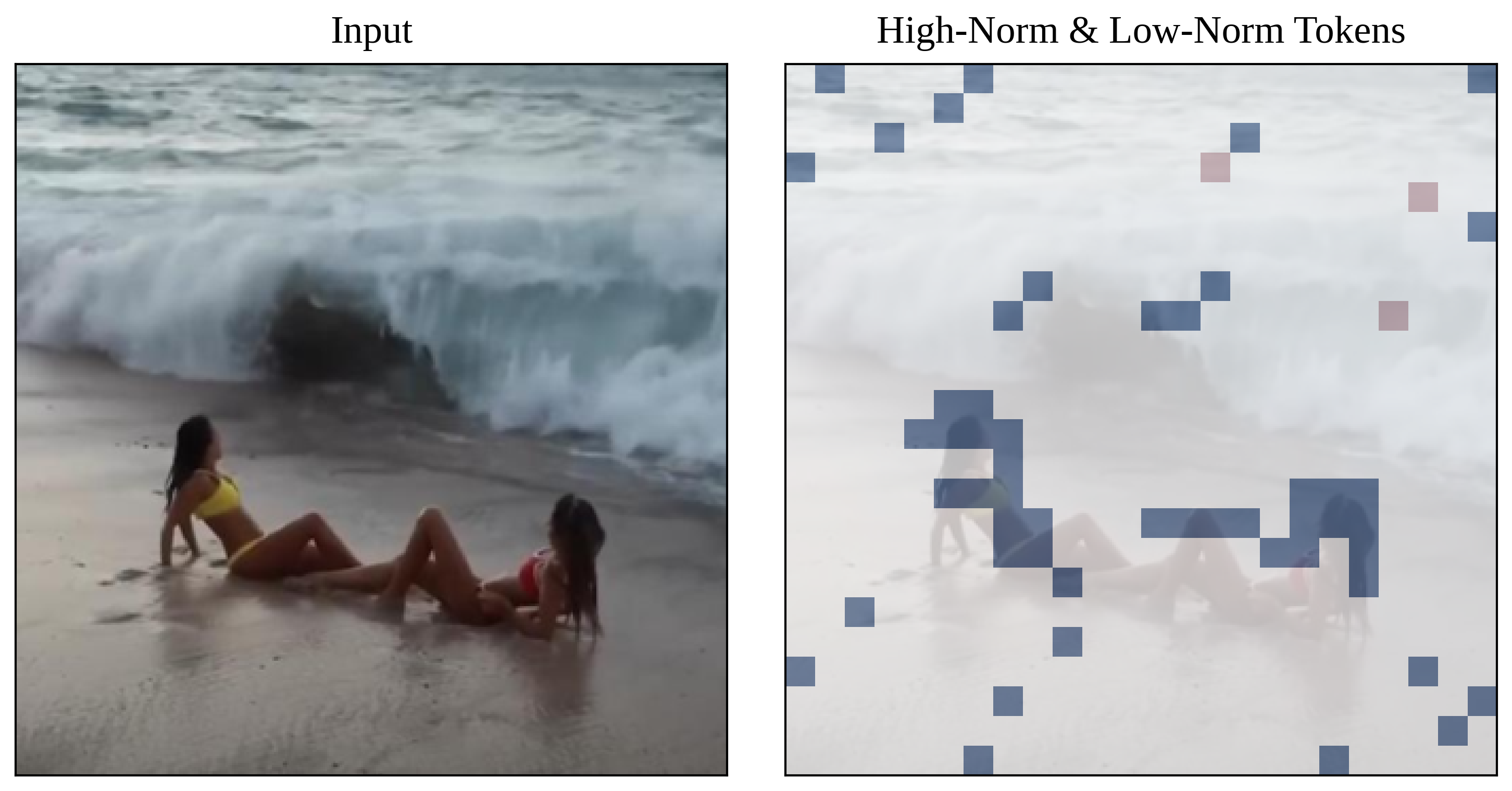}\par\vspace{2pt}
    \includegraphics[width=\linewidth]{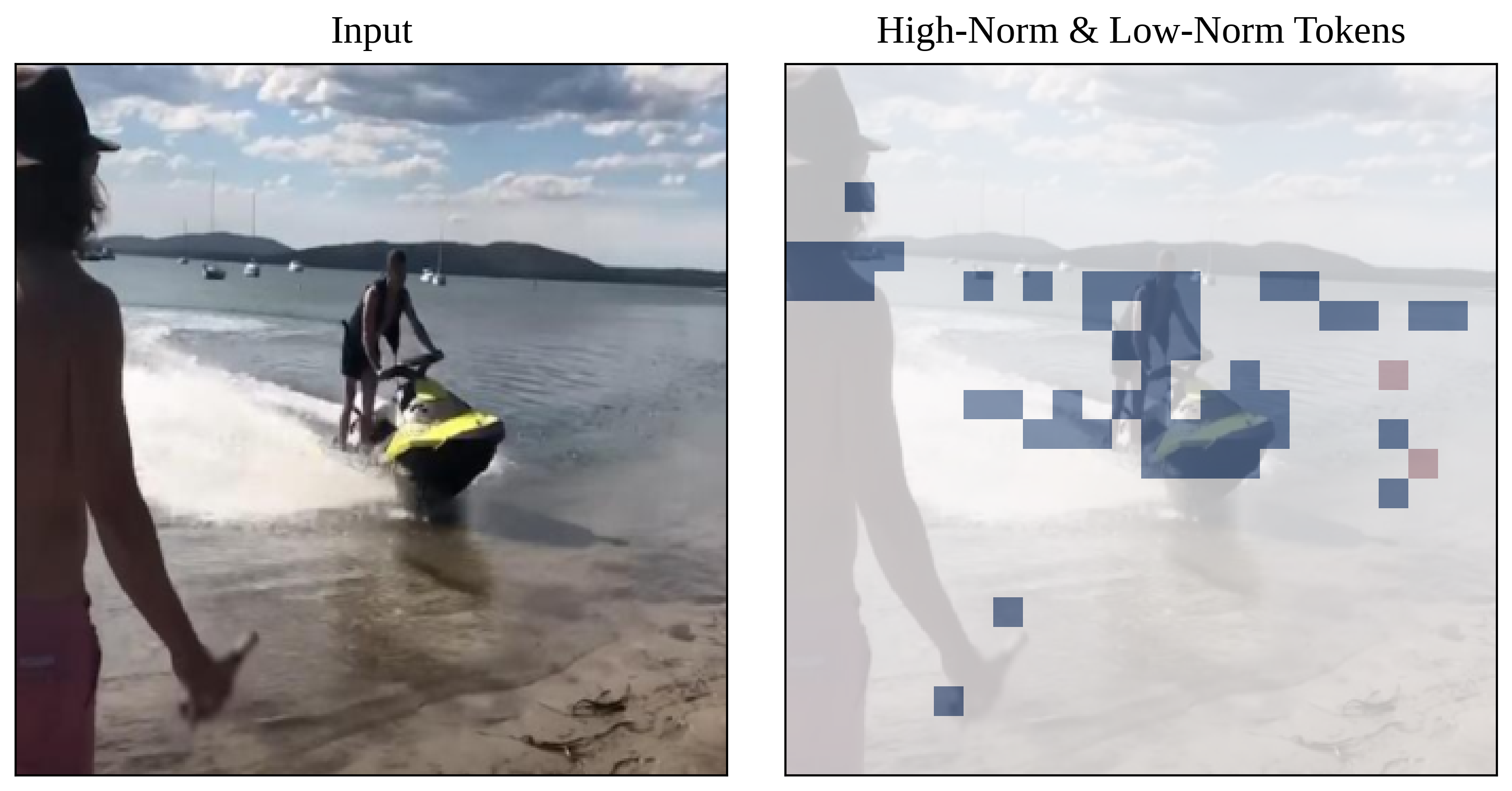}
\end{minipage}

\caption{\textbf{Spatial Redundancy Analysis.}
Visualizations demonstrate that high-norm outlier tokens (red) predominantly appear in non-semantic regions with high local similarity, whereas filtered informative low-norm tokens (blue, top-ranked by \texttt{CLS} attention) correspond to unique local features.}
\label{fig:spatial_redundancy1}
\end{figure*}

Figure~\ref{fig:spatial_redundancy1} further compares the spatial locations of high-norm and informative low-norm tokens across more examples.
We observe that high-norm outliers are mostly concentrated in locally repetitive background areas, while the retained low-norm tokens tend to lie on visually distinctive and semantically meaningful regions.
These visualizations provide additional qualitative support for our claim that high-norm tokens are highly redundant in both spatial and representational dimensions.
See Fig.~\ref{fig:spatial_redundancy1}.

\end{document}